\documentclass{article}

\usepackage[nonatbib,preprint]{neurips_data_2024}
\usepackage[style=ieee, sorting=none, backend=biber, maxbibnames=99, minnames=1, maxnames=10, citestyle=numeric-comp]{biblatex}
\bibliography{bib_neurips}

\usepackage[utf8]{inputenc} 
\usepackage[T1]{fontenc}    
\usepackage{hyperref}       
\usepackage{url}            
\usepackage{booktabs}       
\usepackage{amsfonts}       
\usepackage{nicefrac}       
\usepackage{microtype}      
\usepackage{xcolor}         

\usepackage{booktabs}
\usepackage{multirow}
\usepackage{leftidx}
\usepackage{bbm}
\usepackage{floatrow}

\usepackage{algorithmic}
\usepackage{textcomp}
\usepackage{siunitx}
\usepackage{tablefootnote}
\usepackage{footnote}
\usepackage{caption}
\usepackage{makecell}
\usepackage{pifont}

\usepackage{xspace}
\usepackage{amsmath}
\usepackage{amssymb}
\usepackage{graphicx}
\usepackage{floatrow}
\usepackage{wrapfig}
\newfloatcommand{capbtabbox}{table}[][\FBwidth]

\newcolumntype{P}[1]{>{\centering\arraybackslash}p{#1}}
\newcolumntype{M}[1]{>{\centering\arraybackslash}m{#1}}
\newcolumntype{L}[1]{>{\raggedright\arraybackslash}m{#1}}

\newcommand{\myparagraph}[1]{\vspace{2pt}\noindent{\bf #1}}

\newcommand{\OurMethod}{CaD-VI~}
\newcommand{\OurModelPhaseOne}{CaD-LLaVA$^{V1}$~}
\newcommand{\OurModelPhaseTwo}{CaD-LLaVA$^{V2}$~}
\newcommand{\OurData}{CaD-Inst~}
\newcommand{\OurDataPhaseOne}{CaD-Inst$^{V1}$~}
\newcommand{\OurDataPhaseTwo}{CaD-Inst$^{V2}$~}
\newcommand{\OurApproach}{CaD-VI~}
\newcommand{\CaD}{CaD~}
\newcommand{\CaDfull}{Commonalities and Differences~}

\newcommand{\OurBenchmark}{CaD-QA~}

\makeatletter
\DeclareRobustCommand\onedot{\futurelet\@let@token\@onedot}
\def\@onedot{\ifx\@let@token.\else.\null\fi\xspace}

\def\eg{\emph{e.g}\onedot} 
\def\ie{\emph{i.e}\onedot}

\makeatother

\makeatletter
\def\blfootnote{\gdef\@thefnmark{}\@footnotetext}
\makeatother

\title{Comparison Visual Instruction Tuning}

\author{
Wei Lin$^{\dagger 1}$\and
Muhammad Jehanzeb Mirza$^{2}$ \and 
Sivan Doveh$^{3,4}$ \and 
Rogerio Feris$^7$ \and 
Raja Giryes$^5$ \and 
Sepp Hochreiter$^{1,6}$ \and
Leonid Karlinsky$^7$ \and\\
$^1$ELLIS Unit, LIT AI Lab, Institute for Machine Learning, JKU Linz, Austria\\
$^2$TU Graz ICG, Austria
$^3$IBM Research, Israel
$^4$Weizmann Institute of Science, Israel\\
$^5$Tel-Aviv University, Israel
$^6$NXAI GmbH, Austria
$^7$MIT-IBM Watson AI Lab, USA \\~\\
Project Page: \url{https://wlin-at.github.io/cad_vi}\\
Dataset Repo: \url{https://huggingface.co/datasets/wlin21at/CaD-Inst}
}

\begin{document}
\blfootnote{
{
$\dagger$ Correspondence: \tt\small{wlin2021at@gmail.com}}
} 
\maketitle

\begin{abstract}

  Comparing two images in terms of \CaDfull (\CaD) is a fundamental human capability that forms the basis of advanced visual reasoning and interpretation. It is essential for the generation of detailed and contextually relevant descriptions, performing comparative analysis, novelty detection, and making informed decisions based on visual data. However, surprisingly, little attention has been given to these fundamental concepts in the best current mimic of human visual intelligence - Large Multimodal Models (LMMs). We develop and contribute a new two-phase approach \OurApproach for 
  collecting synthetic visual instructions, together with an instruction-following dataset \OurData containing 349K image pairs with \CaD instructions collected using \OurApproach.  Our approach significantly improves the CaD spotting capabilities in LMMs, advancing the SOTA on a diverse set of related tasks by up to 17.5\%. It is also complementary to existing difference-only instruction datasets, allowing automatic targeted refinement of those resources increasing their effectiveness for CaD tuning by up to 10\%. Additionally, we propose an evaluation benchmark with 7.5K open-ended QAs to assess the \CaD understanding abilities of LMMs. 
\end{abstract}



\vspace{-0.2cm}
\section{Introduction}
\vspace{-0.2cm}



\begin{figure*}[ht]
\vspace{-0.35cm}
\includegraphics[width=\columnwidth]{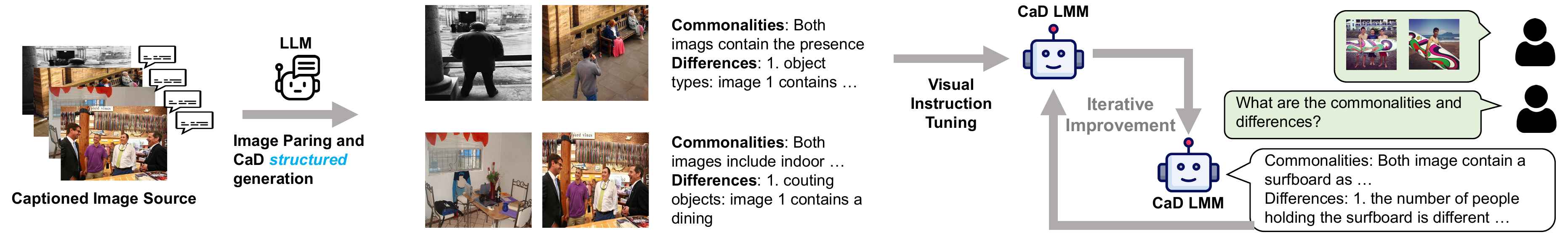}
\caption{
\OurApproach concept. We collect and pair densely captioned source images to form synthetic \CaD instructions using an LLM. The resulting synthetic \CaD Visual Instruction dataset is used to train the first \CaD enabled LMM that is in turn used in iterative self-refinement by annotating new paired images from additional sources using the \CaD LMM, and re-training the model with a growing and more comprehensive \OurData dataset (contributed in this work).
\vspace{-0.15cm}
}\label{fig:concept}
\vspace{-0.1cm}
\end{figure*}

\begin{figure*}[ht]
\vspace{-0.3cm}
\includegraphics[width=\columnwidth]{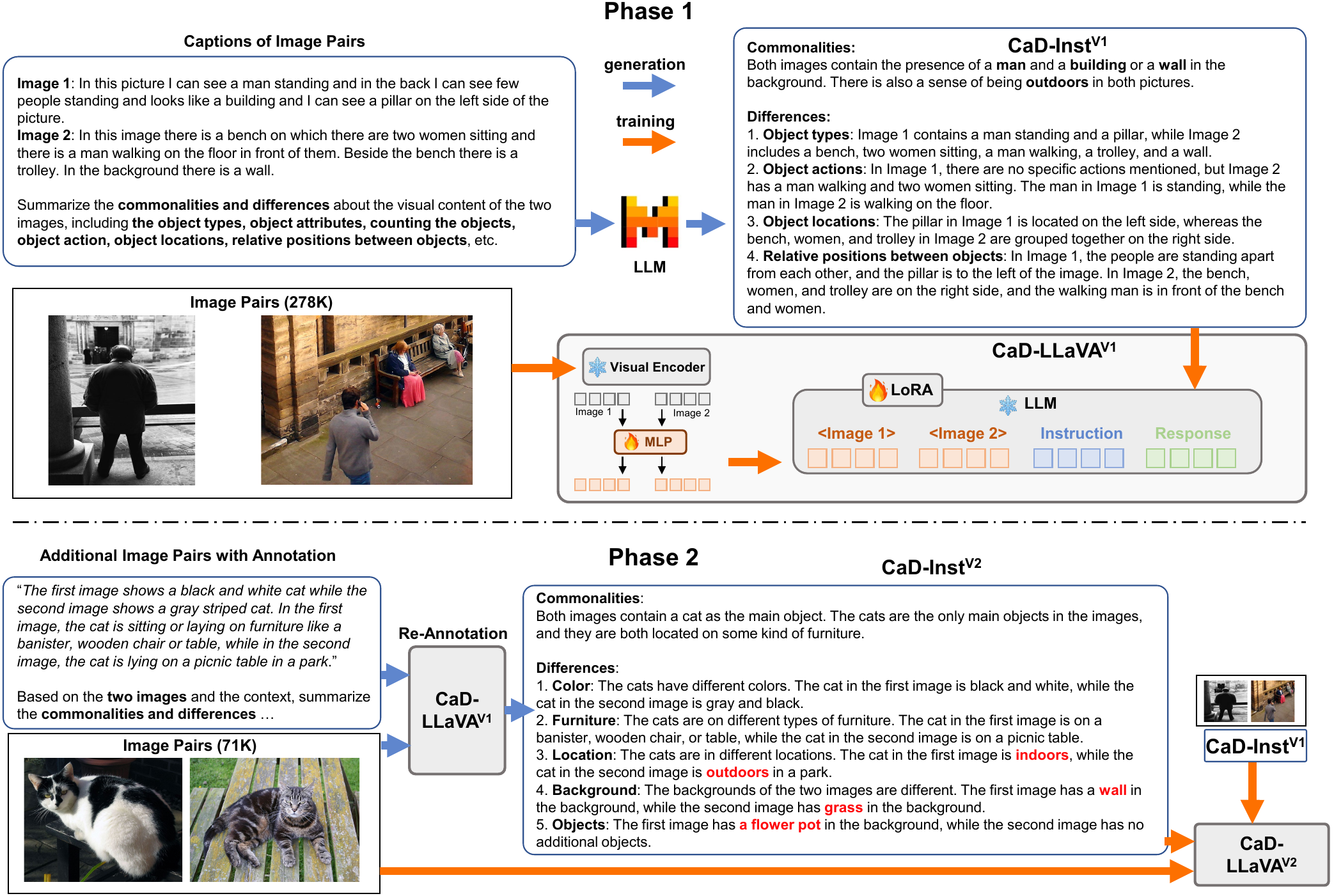}
\caption{
\label{fig:pipeline}
Pipeline of our two-phase \OurApproach: In Phase-1, we leverage captions for image pairs and an LLM to generate \CaD VI data - \OurDataPhaseOne (278K), and perform visual instruction tuning on it to arrive at the Phase-1 model \OurModelPhaseOne. In Phase-2, we leverage \OurModelPhaseOne to generate \CaD VI data on additional image pairs and collect \OurDataPhaseTwo (71K). Visual instruction tuning with \OurDataPhaseOne and \OurDataPhaseTwo leads to our final model \OurModelPhaseTwo. 
\vspace{-0.5cm}
}
\end{figure*}

Understanding the \CaDfull (\CaD) between two signals (e.g., images) is a basic capability innate to humans \cite{gestalt}. Spotting change and difference alerts us to interesting events happening in our surroundings, warns us of hazard, and drives us toward learning new concepts exposed after the change or relative movement. Understanding what is common helps structure visual information and allows differences to emerge by elimination. Together, these form powerful tools for human learning and acquiring world knowledge. 

The forefront of modern AI shifted with the recent emergence of foundation Large Language Models (LLMs) \cite{bommasani2022opportunities}, where the top-performing ones \cite{openai2024gpt4,geminiteam2024gemini,claude,llama3} closely align to human reasoning and world-knowledge capabilities. LLMs' great performance and wide applicability quickly led to their wide adoption into most of the current ML pipelines. In the Vision community, this impacted the development of Large Multi-modal Models (LMMs) \cite{llava,yang2023dawn,geminiteam2024gemini,huang2023sparkles,li2023otter,internlmxcomposer2,Emu2} largely considered the best available mimic of human visual intelligence to date. While multiple methods for adding multi-modal support to LLMs have been proposed, currently the more popular and better performing open LMMs largely rely on tuning using Visual Instructions (VI) \cite{llava,zhu2023minigpt}. These methods align image tokens produced by visual encoders to be `understandable' by an LLM decoder, allowing images to be seamlessly integrated into the LLM decoder input context stream together with the query text during inference. In most recent methods \cite{llava,huang2023sparkles,li2023otter,internlmxcomposer2}, VI takes the form of a multi-turn conversation: with `human' turns providing image context and asking the questions, and LMM turns answering them \cite{llava}. However, the majority of VI data focused on providing merely a single image in the VI conversations~\cite{llava}, while only a few works included multi-image VI samples \cite{Emu2,awadalla2023openflamingo}, and surprisingly, very few included some form of \CaD VI data \cite{huang2023sparkles,li2023otter,li2023mimic} to enable \CaD support in the resulting LMM.

Due to the fundamental importance of endowing LMMs with \CaD capabilities, thus getting them closer to achieving human visual intelligence in all its diversity, we propose \OurApproach - a multi-phase \CaD generation approach, for progressive dense and structured \CaD VI data collection (concept shown in Fig.~\ref{fig:concept}), which we employ to build \OurData training curriculum and associated \OurBenchmark benchmark comprised of CaD-related open-ended questions, both contributed in this work. In essence, the final \OurData curriculum associates diverse and large-scale (349K) image pair collection with highly detailed and structured \CaD summaries. \CaD summaries computed for an additional set of 7.6K image pairs, are used for extracting open CaD-related QA resulting in \OurBenchmark. 

As shown in Fig.~\ref{fig:pipeline}, the Phase-1 of \OurApproach is a `cold start' where, in the absence of LMMs with substantial \CaD capabilities, we leverage image captions and an LLM to hallucinate (coarse) \CaD VI data - \OurDataPhaseOne (278K), where we collect \textit{structured} and \textit{detailed} \CaD summaries for our paired images sourced from a dense \& large-scale image collection~\cite{pont2020connecting}.
%
Training on the first phase \OurDataPhaseOne data we 
arrive at \OurModelPhaseOne - an LMM that has strong \CaD capabilities compared to a large variety of leading LMMs including the very few trained with some \CaD data (see Sec. \ref{sec:results}).  Next, leveraging our \OurModelPhaseOne model to produce non-hallucinated, image-informed \CaD data, we generate additional \CaD instructions into the collection \OurDataPhaseTwo (71K). 
Combining \OurDataPhaseOne and \OurDataPhaseTwo we form \OurData and train our final \OurModelPhaseTwo 7B and 13B LMMs to achieve (1) significant (up to 17.5\%) absolute improvement over a large variety of recent SOTA LMMs over a variety of 5 CaD-related existing closed-QA evaluation benchmarks (namely BISON\cite{hu2019evaluating}, SVO Probes\cite{hendricks2021probing}, NLVR2\cite{suhr2019corpus}, EQBEN\cite{wang2023equivariant}, and COLA\cite{ray2023cola}), and (2) strong (up to over 20\%) relative improvements on our contributed open-QA CaD benchmark - \OurBenchmark. Additionally, as \OurData can be safely mixed with the LLaVA VI data~\cite{llava1_5}, we show in Tab. \ref{tab:eval_seed_img} that our \OurModelPhaseTwo models effectively avoid forgetting the general capabilities of the corresponding LLaVA LMMs.

Our contributions are as follows: (i) we contribute \OurData - a large-scale visual instruction tuning dataset for enhancing \CaD reasoning capabilities of LMMs; (ii) we contribute \OurBenchmark - an open QA evaluation benchmark for assessing \CaD capabilities; (iii) we contribute and open source a \OurApproach methodology for collecting and enhancing \CaD instruction tuning data; (iv) we demonstrate significant (up to 17.5\%) improvements in \CaD reasoning for LMMs trained using \OurData as well as potential to scale \OurData via self-improvement by \OurData-trained models.

\vspace{-0.2cm}
\section{Related Work}
\vspace{-0.2cm}







\myparagraph{Large Multimodal Models.} LMMs have shown significant advancements in integrating visual and textual data, enhancing the ability of deep neural networks to understand and generate multimodal content. BLIP-2 employs a bootstrapping approach that leverages frozen image encoders and large language models through a querying transformer, achieving remarkable results on various vision-language tasks with fewer parameters compared to previous models \cite{li2023blip}.  Similarly, MiniGPT-4 \cite{minigpt4} and LLaMA-Adapters \cite{zhang2023llama} utilize pretrained visual and language models, with adapters aligning image tokens to language tokens, improving the efficiency and performance of multimodal understanding and generation.
In addition to these early models, the LLaVA series \cite{llava}, including LLaVA 1.5 \cite{llava1_5} and LLaVA 1.6 \cite{liu2024llavanext}, have enhanced visual instruction tuning, enabling better handling of single-image inputs and more accurate multimodal outputs. 
The InternLM XComposer 2.0 VL \cite{zhang2023internlm}, EMU2 \cite{sun2023generative}, Otter \cite{li2023otter}, SparklesChat \cite{huang2023sparkles}, and MMICL~\cite{zhao2024mmicl} extend these capabilities by incorporating multiple images as input, thereby enriching the models' understanding and generation of text based on complex visual scenes. These models showcase the evolution from single-image to multi-image inputs, highlighting the progress in multimodal learning architectures and applications.

\myparagraph{Visual Instruction Tuning Datasets.} The success of LMMs builds on the collection of high-quality visual instruction tuning data, either constructed from existing VQA datasets~\cite{gong2023multimodal,goyal2017making,hudson2019gqa,instructblip,li2023m}, curated image-text pairs~\cite{minigpt4} and LLM-generated instruction-following data with input of rich human annotations~\cite{llava,llava1_5,zhang2023llavar,zhao2023svit,li2023mimic}. However, the collection of multimodal data for learning commonalities and differences between two images is still under-explored.

\myparagraph{Image Commonalities and Differences.} 
Only a few datasets contain difference-only related annotation \cite{jhamtani2018learning,li2023mimic}.
Spot-the-diff~\cite{spotthediff} collects human-annotated short change descriptions for surveillance video frames. 
%
Our \OurDataPhaseOne data collection is partially inspired by the differences-only data collection done by \cite{li2023mimic} as a small part of their VI strategy. However, different from \cite{li2023mimic} we: (i) collect both differences \textit{and commonalities} (compared to only differences in \cite{li2023mimic}); (ii) we leverage a significantly more \textit{dense} caption-source of \cite{pont2020connecting} compared to \cite{chen2015microsoft} used in 
\cite{li2023mimic}; (iii) we are \textit{structuring} our differences in \CaD according to 6 axes (whichever applicable on case basis) - object types, attributes, counting, actions, locations, and relative positioning, also explicitly asking the LLM to extract (from the dense captions) information along these axes, while \cite{li2023mimic} produced unstructured difference description text; (iv) unlike \cite{li2023mimic} we are not relying on the existence of manually collected object bounding boxes; (v) the scale of our data is approx. 4 times larger than of \cite{li2023mimic}. 
Due to these differences, as evident from the direct comparison in Tab. \ref{tab:train_on_scene_diff}, training the same model on \OurDataPhaseOne has significant performance advantages over training on CaD instructions of \cite{li2023mimic}.
To summarize, our work focuses on \CaD understanding, largely neglected by the visual instruction tuning community. We propose a new \OurApproach approach for collecting synthetic visual instructions and enhancing the \CaD analysis capabilities in LMMs. \OurApproach not only advances the state-of-the-art in related tasks by significant margins but also complements existing datasets  \cite{jhamtani2018learning,li2023mimic} by enabling their automatic targeted refinement, thereby improving their effectiveness for \CaD tuning.

\vspace{-0.2cm}
\section{\OurMethod - Two-Phase \CaD Visual Instruction Tuning}
\vspace{-0.2cm}

As illustrated in Fig.~\ref{fig:pipeline}, our \OurMethod consists of two phases: in Phase-1, we employ an LLM to generate summary of CaD for image pairs (Sec.~\ref{sec:llm_instruct_data_collect}) and perform visual instruction tuning on the collected data (Sec.~\ref{sec:visual_instruct_tune}); in Phase-2, we leverage the Phase-1 model to generate CaD on additional image pairs and perform training with combined instruction data from both phases (Sec.~\ref{sec:phase_2_data_collect_instruct_tune}). 

\vspace{-0.2cm}
\subsection{Phase-1a: LLM Instruction Data Collection - \OurDataPhaseOne}\label{sec:llm_instruct_data_collect}
\vspace{-0.2cm}
In our first phase, we leverage an LLM to generate a summary of commonalities and differences for a pair of two images, as shown in Fig.~\ref{fig:pipeline} (top row). Specifically, we construct image pairs and prompt an LLM, supplying it with two image captions (one per image) and an instruction prompt asking it to summarize all the commonalities and differences according to the provided captions, contributing to our first phase \CaD instruction data collection denoted as \OurDataPhaseOne.

\myparagraph{Image Source.} 
We select the Localized Narratives dataset~\cite{pont2020connecting} which consists of 873K image-caption pairs with diverse samples sourced from COCO~\cite{lin2014microsoft,chen2015microsoft}, Flickr30K~\cite{young2014image}, ADE20K~\cite{zhou2019semantic} and Open Images~\cite{kuznetsova2020open}. The captions are generated by transcription from spoken descriptions of the image content, which are quite dense, detailed, and descriptive with an average length of 36.5 words.  
To cover comprehensive visual contents and increase the diversity in terms of commonalities and differences, we collect 278K image pairs with different levels of similarity between their captions. 
We compute similarity by counting the number of overlapping nouns in the corresponding captions.


\begin{figure*}[ht]
\includegraphics[width=\columnwidth]{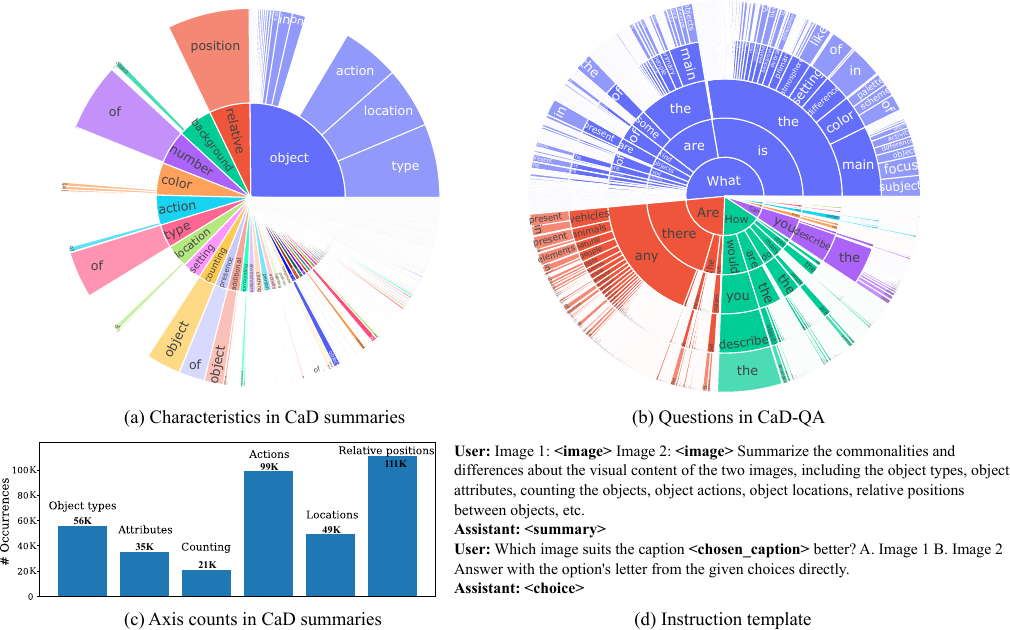}
\caption{
\label{fig:statistics_combined}
(a) Distribution of characteristics (first two words) in the CaD summary collected in \OurDataPhaseOne; 
 (b) Distribution of question types (first five words) in the evaluation benchmark \OurBenchmark;
 (c) Axis counts in CaD summaries;
 (d) Two-turn conversation template.
 \vspace{-0.2cm}
}
\vspace{-0.2cm}
\end{figure*}

\myparagraph{LLM Data Generation.} 
Inspired by LLaVA~\cite{llava} who used an LLM for single images visual instruction collection, we leverage the Mixtral 8$\times$7B open LLM~\cite{jiang2024mixtral} for generating detailed and structured summaries of commonalities and differences for pairs of images. As the LLM can only accept text as input, in Phase 1 we use image captions to represent visual content of images. This is a rather crude approximation, which is alleviated in Phase 2 of our \OurApproach approach. 
To encourage the diverse and creative generation of commonalities and differences, we do not provide in-context examples of expected output in the prompt to the LLM. Furthermore, we specifically prompt the LLM \textit{to structure} the commonalities and differences summaries according to the following 6 visual aspects: (i) object types; (ii) attributes; (iii) counts; (iv) actions; (v) locations; and (vi) relative positions; as illustrated in Fig.~\ref{fig:pipeline}. We provide detailed prompts in the Supplementary. Importantly, LLM is not forced to produce all 6 aspects in every summary; 
they are generated adaptively according to the available content.

\myparagraph{Generated Data Statistics.}
In \OurDataPhaseOne we collected structured summaries of \CaD for 278K image pairs, with average length of 157 words (40 for commonalities and 117 for differences). The summaries are structured according to 6 axes, appearing unevenly on a case-to-case basis based on the LLM decision. We illustrate the distribution of data characteristics in Fig. \ref{fig:statistics_combined}(a), and the total observed axis counts in Fig.~\ref{fig:statistics_combined}(c). More statistics and details are provided in the Supplementary.

\myparagraph{\CaD visual instructions data.} 
We construct a two-turn conversation for each image pair. In the first turn, we define the task of summarizing \CaD by providing the encoded visual tokens of the two images and instructing the model to summarize the \CaD, where the response part of the turn is the LLM-generated structured summary collected above. In this instruction, we do not provide the image captions, forcing the model to rely only on image tokens to complete the task. In the second turn, we reinforce the image-text alignment by employing a simple task of text-to-image retrieval to avoid forgetting the model's general capabilities. We randomly sample one of the two captions and request the model to select the image (from the current pair) to which the caption belongs.
The template for the two-turn conversation is illustrated in Fig.~\ref{fig:statistics_combined}(d).  

\vspace{-0.2cm}
\subsection{Phase-1b: CaD Visual Instruction Tuning}\label{sec:visual_instruct_tune}
\vspace{-0.2cm}
\myparagraph{Architecture.} As illustrated in Fig.~\ref{fig:pipeline}, we use our collected \OurDataPhaseOne data to perform visual instruction tuning using the open-sourced code of LLaVA-1.5~\cite{llava1_5} LMM. The LLaVA-1.5 model consists of $\phi_L(\cdot;\theta_L)$ - a pretrained Vicuna 1.5~\cite{zheng2023judging} LLM (finetuned from LLama 2~\cite{llama2}); $\phi_V(\cdot;\theta_V)$ - a pretrained visual encoder CLIP ViT-L/14@336px~\cite{clip}; and $\phi_M(\cdot;\theta_M)$ - a two-layer MLP projector converting the visual encoder tokens to post-embedding layer LLM tokens. 

Given a pair of two images $x_{V_1}$, $x_{V_2}$ and the instruction $x_I$, the MLP projects the visual features computed by the visual encoder into embedded language tokens, \ie $v_k = \phi_M(\phi_V(x_{V_k};\theta_V);\theta_M), k\in\{1,2\}$. Then the projected visual features and instruction text tokens are concatenated and fed into the LLM, where the response text tokens are generated in an autoregressive manner, \ie
\begin{equation}
    \hat x^i_R = \phi_L([ v_1, v_2, x_I, \hat x^{<i}_R  ]  ;\theta_L), 
\end{equation}
where $\hat x^i_R$ denotes the $i$-th token in the generated response. 

\myparagraph{Training.} We finetune the LLaVA-1.5 model using the LLaVA~\cite{llava} pipeline. Specifically, following LLaVA pre-training, 
we finetune only the pretrained projection MLP and the (frozen) LLM with LoRA adapters~\cite{hu2021lora}. 
We minimize the CLM loss of the next token prediction in the responses:
\begin{equation}
    \mathcal{L}_{CLM} = \sum_i -\log p( {\hat x}^i_{R} | {V_1}, {V_2}, {x_I},  
    {x^{<i}_R} ) 
\end{equation}
To preserve the general VL capabilities of the LMM, we merge our \OurDataPhaseOne with the finetuning data of LLaVA-1.5 (665K samples). In Tab. \ref{tab:eval_seed_img} we show that \OurMethod indeed preserves the general LMM capabilities compared to LLaVA-1.5 as evaluated on the popular SEED benchmark \cite{li2023seed}.
The Phase-1 CaD visual instruction tuning results in our cold-start model \OurModelPhaseOne which is an LMM that can be leveraged for annotating visual commonalities and differences.  

\vspace{-0.2cm}
\subsection{Phase-2: Data Collection and Visual Instruction Tuning}\label{sec:phase_2_data_collect_instruct_tune}
\vspace{-0.2cm}

\myparagraph{Phase-2a: LMM-based \CaD Instruction Collection.}
While in Phase 1 we used an LLM to extract a \CaD summary based on human-generated captions, for Phase 2 data collection we leverage our Phase 1 model \OurModelPhaseOne and additional image pairs to extract the \CaD summaries informed by the images directly. 
Here we select the Scene-Difference~\cite{li2023mimic} collection as an additional image source. It contains 71K pairs of similar images from COCO~\cite{lin2014microsoft} and provides annotation of unstructured difference-only summaries (see Fig.~\ref{fig:pipeline} bottom left for an example). We feed both the image pairs and the original annotations into our \OurModelPhaseOne model, and generate a \textit{structured summary} of \textit{both} commonalities and differences. The exact prompt is provided in the Supplementary. This leads to our phase-2 \CaD instruction data - \OurDataPhaseTwo. As shown in Tab. \ref{tab:train_on_scene_diff}, our collected \CaD instructions significantly improve over the utility of the original \cite{li2023mimic} annotations. As part of our analysis in Tab. \ref{tab:train_on_scene_diff} and \ref{tab:train_on_eqben}, and additional experiments provided in Supplementary, we also show that similarly out-of-distribution image pair collections or even unlabeled image pair collections can be effectively leveraged for our Phase-2.

\myparagraph{Phase-2b CaD Visual Instruction Tuning}
We follow the Phase-1b introduced in Sec.~\ref{sec:visual_instruct_tune} for CaD visual instruction tuning. Here we finetune on a combination of LLaVA 1.5 \cite{llava1_5} finetune data (665K), \OurDataPhaseOne data (278K) and \OurDataPhaseTwo data (71K). This phase of \CaD visual instruction tuning leads to the Phase 2 model, denoted as \OurModelPhaseTwo.

\vspace{-0.2cm}
\section{\OurBenchmark - Benchmark of Open-Ended \CaD QA}\label{sec:benchmark_collection}
\label{sec:bench}
\vspace{-0.2cm}



In order to evaluate the capability of LMMs on answering open-ended questions regarding commonalities and differences of a pair of two images, we construct and contribute the \OurBenchmark benchmark. 

\myparagraph{Data Collection.} 
Similar to the data collection pipeline introduced in Sec.~\ref{sec:llm_instruct_data_collect}, 
we employ Visual Genome~\cite{krishna2017visual} and the detailed image captions from SVIT~\cite{zhao2023svit} as image \& caption source.  We collect 7.5K image pairs with 8 or more overlapping nouns in their captions. For each pair, we employ the Mixtral 8$\times$7B LLM to produce the structured \CaD summaries from the captions. Next, we prompt Mixtral with both the image captions and the \CaD summary, instructing it to generate a multi-turn conversation with several rounds of Q\&A, providing some in-context examples of the desired layout (see Supplementary for the prompt). Finally, we randomly select one Q\&A per conversation. 

\myparagraph{Benchmark Statistics.} 
There are 7520 QA pairs with an average answer length of 26 words. Among these, we also include 2916 questions asking about the content of only one of the two images. It requires the precise attention of the LMM on the corresponding image to correctly answer these questions. Our \OurBenchmark covers diverse question types as illustrated in Fig.~\ref{fig:statistics_combined}(b). 



\myparagraph{LLM-assisted Evaluation.} Motivated by LLMs' ability to judge response quality consistently with human assessment~\cite{zheng2023judging}, we employ the Mixtral 8$\times$7B LLM to compare the generated responses to the collected open-ended QA responses. We feed the question, correct answer, and the predicted answer into the LLM and instruct it to provide a rating between 0 and 5 for the predicted answer quality (where higher score indicates a better prediction). We provide the prompt in the Supplementary.


\begin{table*} [!tb]
\vspace{-0.4cm}
    \centering
    \scriptsize
    \begin{tabular}{cM{1.6cm}M{1cm}M{1cm}M{1cm}M{0.9cm}M{1.2cm}M{1.2cm}M{0.7cm}cccc}
    \toprule
    Dataset & \multirow{2}*{\makecell{\# Instruction\\Data}} &  BISON &  SVO  & NLVR2 & EQBEN & COLA \\
    Random chance & ~ & 50\% & 50\% & 50\% & 25\% & 25\% \\
    \midrule

    SparklesChat & 6.5K &  56.70\% & 43.93\% & 58.00\% & 19.17\% & 20.00\%  \\
    Otter & 2.8M & 40.67\% & 47.33\% & 52.00\% &  \phantom{0}8.33\% & \phantom{0}8.10\% \\
    MMICL   & 5.8M & 80.00\% & 88.13\%  & 56.67\% & 20.83\%  & 25.71\%  \\
    EMU2-Chat & 1.3M & 46.00\% & 47.93\% & 60.00\% & \phantom{0}7.50\% & 13.33\%   \\
    InternLM-XComposer2-VL & >600K & 80.67\% & 82.07\% & \underline{66.67\%} & 25.00\% & 32.38\%  \\
    LLaVA 1.6 7B  & <1M & 66.00\% & 70.40\% & 58.67\% & 20.83\% & 11.90\% \\
    LLaVA 1.6 13B  & <1M & 81.33\% & 82.13\% & 60.00\% & 17.50\% & 24.76\% \\

    \midrule
    LLaVA 1.5 7B  & 665K & 54.00\% & 46.80\% & 61.33\% & 17.50\% & \phantom{0}7.62\%  \\
    LLaVA 1.5 13B  & 665K & 59.33\% &  56.27\% & 66.00\% & 16.67\% & 12.38\% \\
    \midrule
    \OurMethod 7B & 1M & \underline{95.33\%}  & \underline{92.73\%}  & \underline{66.67\%}  & \underline{39.17\%} & \underline{40.95\%}  \\
    \OurMethod 13B & 1M & \textbf{96.67\%}
    & \textbf{93.00\%} & \textbf{69.33\%} & \textbf{42.50\%} & \textbf{43.33\%} \\
    \bottomrule
    \end{tabular}
    \caption{Performance on closed-ended VQA tasks with image pairs in accuracy. Here the method \OurMethod denotes our Phase-2 model \OurModelPhaseTwo. 
    \vspace{-0.4cm}
    }
    \label{tab:close_end_eval}
\end{table*}

\vspace{-0.2cm}
\section{Experiments}\label{sec:results}
\vspace{-0.2cm}
\myparagraph{Evaluation Datasets}
We evaluate on several VQA benchmarks of closed-ended and open-ended questions. 
For \textbf{closed-ended VQA on image pairs}, we include  BISON~\cite{hu2019evaluating} and SVO Probes~\cite{hendricks2021probing} both consisting of samples with an image pair and a text query that needs to be matched with one of the images in the pair (chance is 50\%). 
EQBEN~\cite{wang2023equivariant} and COLA~\cite{ray2023cola} contain samples composed of a pair of two images together with the two textual descriptions. 
The goal is to correctly match images with corresponding texts (chance is 25\%).
Furthermore, we evaluate on NLVR2~\cite{suhr2019corpus} which comprises samples of a pair of two images and a reasoning sentence. The task is to assess the correctness of the reasoning and has a random chance of 50\%. We also evaluate SEED-Bench Video~\cite{li2023seed} with two frames sampled from the video to explore the generalization value of our \CaD tuning for video understanding. SEED-Bench Video contains three partitions from SEED-Bench and has multi-choice questions on action recognition/prediction or procedure understanding with four answer options per question. 
For \textbf{open-ended tasks}, use the LLM-as-a-judge metric (Sec. \ref{sec:bench}). We evaluate open-ended QAs on our \OurBenchmark. Furthermore, we also directly evaluate the quality of LMM predicted \CaD summaries for 210 image pairs in COLA with shorter summaries generated from brief captions, and for the 7.5K lengthy summaries from \OurBenchmark generated from detailed VG captions. More details and statistics of the datasets are provided in the Supplementary.

\myparagraph{Implementation Details}
We leverage the Mixtral 8$\times7$B Instruct v0.1 and set the maximum token size to 750 data collection and 20 for open-ended task evaluation. For visual instruction tuning, we use the official implementation of LLaVA and tune the LLaVA 1.5 7B model with LoRA. We set the batch size to 128 and LoRA learning rate for LLM and the projector is set to $1\times 10^{-4}$ and $2\times 10^{-5}$ correspondingly. All experiments are run on 4$\times$A100 80G GPUs. More details are in Supplementary.

\begin{minipage}[c]{0.6\textwidth}
\centering
\scriptsize
    \begin{tabular}{cccM{1cm}M{1cm}M{0.7cm}M{0.7cm}M{0.7cm}M{0.7cm}cccc}
    \toprule
    \# Input Frames & 1 & 2 \\
    \midrule

        SparklesChat & 21.81\% &  19.09\%  ($\blacktriangledown$-2.72\%) \\
        Otter & 18.19\% & 23.00\% ($\blacktriangle$\textbf{+4.81\%}) \\
        EMU2-Chat  & \textbf{43.43\%} &  41.09\% ($\blacktriangledown$-2.34\%)  \\
        InternLM-XComposer2-VL  & 41.07\% & 40.16\% ($\blacktriangledown$-0.91\%)\\
        LLaVA 1.6 7B  & \underline{41.95\%} & \underline{42.03\%} ($\blacktriangle$+0.08\%)  \\
        LLaVA 1.6 13B  & 41.85\% & 41.35\% ($\blacktriangledown$-0.50\%)  \\
        \midrule
        LLaVA 1.5 7B   & 37.43\% & 36.68\% ($\blacktriangledown$-0.75\%) \\
        LLaVA 1.5 13B  & 40.12\% & 38.78\% ($\blacktriangledown$-1.34\%) \\
        \midrule
        \OurMethod 7B & 38.40\% & 40.44\% ($\blacktriangle$+2.04\%)  \\
        \OurMethod 13B  & 40.16\% & \textbf{43.09\%} ($\blacktriangle$\underline{+2.93\%}) \\
    \bottomrule
    \end{tabular}
\captionof{table}{Performance on SEED-Bench video partitions by feeding one or two frames into the LMMs.}
\label{tab:eval_seed_vid}
\end{minipage}
\hspace{0.3 in}
\begin{minipage}[c]{0.25\textwidth}
\centering
\scriptsize
    \begin{tabular}{cc}
    \toprule
    Model & SEED-Image   \\
    \midrule
       LLaVA 1.5 7B & 67.34\%  \\
        \OurMethod 7B & 67.48\%  \\
        \midrule
       LLaVA 1.5 13B  & 68.83\% \\
        \OurMethod 13B & 69.11\%  \\

    \bottomrule
    \end{tabular}
\captionof{table}{Performance on SEED-Bench image partitions for evaluation of general VL capabilities with single-image input.}
\label{tab:eval_seed_img}
\end{minipage}

\myparagraph{Comparison to State-of-the-Art LMMs}
\begin{table*} 
\vspace{-0.6cm}
    \centering
    \scriptsize
    \begin{tabular}{M{3.6cm}M{1.3cm}M{1.6cm}M{1.2cm}ccccccccc}
    \toprule
    Dataset &  \OurBenchmark  & VG comm. & VG diff. & COLA comm. & COLA diff. \\
    \midrule

        SparklesChat &  3.01 &  \underline{2.41} & 3.12 &  1.52 &  1.22 \\
        Otter & 2.20 & 1.88 & 1.97 & 1.37 & 0.81 \\
        MMICL  & 2.01 & 1.79 & 1.94 & 1.73 & 0.59 \\
        EMU2-Chat & 1.20 & 1.04 & 1.08 &  1.22 &  0.41 \\
        InternLM-XComposer2-VL & 2.90 & 2.08 & 2.69 &  1.72 &  \textbf{1.36} \\
        LLaVA 1.6 7B &  3.10 & 2.23 & 2.73 & 1.71 &  1.22\\
        LLaVA 1.6 13B & 3.19 &  2.19 & 2.69 & 1.93 &  1.01 \\
        
        \midrule
        LLaVA 1.5 7B & 2.54 & 1.79 & 1.75 &  1.44 & 1.02 \\
        LLaVA 1.5 13B &  2.65 & 2.16 &  2.41 &  1.57 &  1.10 \\
        \midrule
        \OurMethod 7B &  \underline{3.29} & 2.32 &  \textbf{3.85} &  \textbf{2.14} &  1.25 \\
        \OurMethod 13B &  \textbf{3.34} & \textbf{2.58} &  \underline{3.68} &  \underline{2.13} &  \underline{1.31} \\

    \bottomrule
    \end{tabular}
    \caption{ Performance on \OurBenchmark and tasks of \CaD summary prediction evaluated using LLM-as-a-judge ratings (range 0 to 5). Here the method \OurMethod denotes our Phase-2 model \OurModelPhaseTwo. 
    \vspace{-0.6cm}
    }
    \label{tab:open_ended_eval}
\end{table*}
We first compare our final model \OurModelPhaseTwo (denoted by \OurMethod in Table) to state-of-the-art LMMs on closed-ended VQA in Table~\ref{tab:close_end_eval}. SparklesChat \cite{huang2023sparkles}, Otter \cite{li2023otter}, MMICL \cite{zhao2024mmicl}, EMU2-Chat \cite{Emu2}, InternLM-Xcomposer2-VL \cite{zhang2023internlm} all include samples with multi-image inputs in the visual instruction tuning while LLaVA 1.5 \cite{llava1_5} and LLaVA 1.6 \cite{liu2024llavanext} are tuned with only single image instructions.  The evaluated benchmarks are challenging due to the visually very similar image pairs with subtle compositional differences where the LMMs could easily make an incorrect decision leading to performance below random chance. 
Our \OurMethod 7B model already outperforms all the other baselines on the five benchmarks and our 13B finetuned model further boosts the performance. 

Table~\ref{tab:open_ended_eval} demonstrates the comparison to the baseline LMMs on open-ended tasks of \OurBenchmark and of \CaD summary prediction on image pairs. 
Our \OurMethod models outperform the baselines on four of the five open-ended tasks, with the exception of COLA difference summary where our 13B model achieves a rating (1.31) close to the best performing InternLM-XComposer2 model (1.36).

Furthermore, we explore whether our \CaD instruction tuning improves video understanding evaluated using SEED-Bench Video in Table~\ref{tab:eval_seed_vid}. In the evaluation setting of LLaVA, only one frame per SEED-Bench video is passed to the LMM. To explore the impact of our \CaD tuning, we compare this to evaluating using two frames as input. As shown in Table~\ref{tab:eval_seed_vid}, although multiple baseline LMMs achieve better performance in single-frame setting, our \OurMethod 13B model performs the best in the two-frame setting with a significant performance improvement of 2.93\% on top of the single-frame performance. The only higher improvement is achieved by Otter, which however struggles below the 25\% chance level performance. This underlines that our \CaD tuning improves the temporal understanding between video frames. 

Additionally, to verify that introducing multi-image \CaD data into the tuning does not lead to catastrophic forgetting of general single-image input LMM capabilities, we also evaluate the SEED-Bench Image partitions and report the results in Table~\ref{tab:eval_seed_img}. Here we directly compare to same architecture baseline of LLaVA 1.5 fine-tuned using its single-image LLaVA mix 665K data. 
Table~\ref{tab:eval_seed_img} demonstrates that 
our \CaD tuning indeed preserves the competence in single-image understanding.






\begin{table*} [!htb]
    \centering
    \scriptsize
    \begin{tabular}{M{0.05cm}M{6.2cm}M{0.8cm}M{0.8cm}M{0.8cm}M{0.8cm}M{1.3cm}cccc}
    \toprule
    & Training Data &   BISON &  SVO  &  EQBEN & COLA & \OurBenchmark \\
    \midrule

    A:& LLaVA mix (L)   & 54.00\% & 46.80\% &  17.50\% & \phantom{0}7.62\%  & 2.54\\

     \midrule
   B:& L + ScDiff orig. annot.  & 92.67\% & 90.07\% &  22.50\% & 33.81\% & 2.90\\
   C:& L + ScDiff our annot. (from scratch) & 88.67\% & 90.80\% &  \underline{38.33\%} & \underline{36.67\%} & 3.17 \\
   D:& \makecell{ L + ScDiff our annot.\\(refined from orig. annot.) } & \underline{94.67\%} & 91.80\% &  32.50\% & 34.76\% & 3.17 \\
   \midrule
   E:& \makecell{ L + \OurDataPhaseOne  } & 92.00\% & \underline{92.27\%} &  34.17\%  & \underline{36.67\%} & \underline{3.27}\\
   F:& L + \OurDataPhaseOne + ScDiff our annot. (refined from orig. annot.)  & \textbf{95.33\%} & \textbf{92.73\%} & \textbf{39.17\%} & \textbf{40.95\%} & \textbf{3.29} \\

    \bottomrule
    \end{tabular}
    \caption{
    Ablation of phase-2 data collection from 71K image pairs in Scene-Difference (ScDiff). 
    We use \OurModelPhaseOne to generate CaD on ScDiff either from scratch or by refining from the original annotation of unstructured difference-only summaries. Training settings in E and F lead to our \OurModelPhaseOne and \OurModelPhaseTwo models correspondingly.
    }
    \label{tab:train_on_scene_diff}
\end{table*}

\begin{table*} [!tb]
    \centering
    \scriptsize
    \begin{tabular}{M{0.05cm}M{6cm}M{0.9cm}M{0.9cm}M{0.9cm}M{0.9cm}M{1.3cm}cccc}
    \toprule
    & Training Data  &  BISON &  SVO  &  EQBEN & COLA & \OurBenchmark \\
    \midrule
    A:& LLaVA mix (L)   & 54.00\% & 46.80\% &  17.50\% & \phantom{0}7.62\%  & 2.54\\
    B:&  L + A/G orig. captions only   & 55.33\% & 55.67\% &  \phantom{0}3.33\% & \phantom{0}2.86\% & 2.78\\
    C:& L + A/G our annot. (from scratch) & \textbf{90.00\%} & \textbf{88.53\%} &  \underline{40.83\%} & \textbf{42.86\%} & \textbf{3.21} \\
    D:& L +  A/G our annot. (given orig. captions)  & \underline{88.00\%} &  \underline{86.87\%} &  \textbf{43.33\%} & \underline{30.48\%} & \underline{3.06}\\

    \bottomrule
    \end{tabular}
    \caption{ 
    Ablation of phase-2 data collection from 66K pairs of video frames in Action Genome and GEBC (A/G). We use \OurModelPhaseOne to generate CaD on A/G either from scratch or with the prior information from the original frame captions. 
     }
    \label{tab:train_on_eqben}
\end{table*}

\vspace{-0.2cm}
\section{Ablations}
\vspace{-0.2cm}
\myparagraph{Phase-2 Data Collection analysis.} 
Our Phase-2 data collection introduced in Sec.~\ref{sec:phase_2_data_collect_instruct_tune} can be used to leverage image pairs from various sources for producing effective \CaD instructions. We first ablate the data collection from the 71K image pairs in Scene-Difference~\cite{li2023mimic} (ScDiff) which contains annotation of unstructured difference-only summaries. As shown in Table~\ref{tab:train_on_scene_diff}, training with original annotation of difference-only summaries (row B) significantly improves on the baseline of training with LLaVA data only (row A). Then we show that using \OurModelPhaseOne to generate \CaD instructions on ScDiff remarkably improves further, either if used from scratch (row C) or by refining from the original annotation (row D, also illustrated in Fig.~\ref{fig:pipeline} bottom row). Training with our re-annotation from scratch outperforms the original annotation on all datasets except for BISON. Our re-annotation by refining the original annotation leads to a more balanced performance improvement and is used as the phase-2 instruction data \OurDataPhaseTwo. We combine this with our phase-1 data \OurDataPhaseOne and demonstrate the further performance boost in row F of Table~\ref{tab:train_on_scene_diff}. 

In order to show the robustness of \CaD data collection capability using our \OurModelPhaseOne model, we also explore applying our phase-2 data collection to visually similar frames from user videos in Action Genome and GEBC (A/G). 
In Table~\ref{tab:train_on_eqben}, we first train a baseline using original frame captions only and a simple instruction task of image description (row B), which leads to a significant performance drop on EQBEN and COLA, and minimal improvement on other datasets. Then we use our \OurModelPhaseOne to generate \CaD instructions on the frame pairs either from scratch (row C) or conditioned on the frame captions (row D). Interestingly, on most datasets \CaD instructions generated by our \OurModelPhaseOne from scratch are found to be more effective than ones generated using original captions conditioning, likely due to lack of detail in these captions. 
This once again demonstrates that our model is effective in generating \CaD instructions on unlabeled data. 



\begin{table*} 
    \centering
    \scriptsize
    \begin{tabular}{M{0.05cm}M{4.7cm}cccccccccccc}
    \toprule
    &  Training Data &  BISON & SVO &  \OurBenchmark & VG comm. & VG diff. \\ 
    \midrule
    A:& LLaVA mix (L)  & 54.00\% &   46.80\% &   2.54 &  1.79 &  1.75 \\
    B:& L + t2i retriev. & 58.00\% & 51.33\% & 2.47 & 1.58 & 1.46   \\
    C:& L + comm.  & 64.67\% & 79.73\%  &  3.23 &  \textbf{2.67} &  2.52  \\
    D:& L + diff. & 55.33\% & 72.13\% &  3.24 &  1.97 &  2.89  \\ 
    E:& L + comm. + diff.  & 72.00\% & 82.60\% &  3.24 &  2.13 &  3.42 \\
    F:& L + comm. + diff. + t2i retriev. & \underline{92.00\%} & \underline{92.27\%} &  \underline{3.27}  &   2.21 & \underline{3.69} \\
    G:& F + \OurDataPhaseTwo & \textbf{95.33\%} & \textbf{92.73\%} &  \textbf{3.29} &   \underline{2.32} & \textbf{3.85}  \\


    \bottomrule
    \end{tabular}
    \caption{Ablation on components in the instruction data. Training settings in F and G lead to our \OurModelPhaseOne and \OurModelPhaseTwo models correspondingly. Here \textit{t2i retriev.} refers to the text-to-image retrieval task (see Sec.~\ref{sec:llm_instruct_data_collect}). Training settings in F and G lead to our \OurModelPhaseOne and \OurModelPhaseTwo models correspondingly. 
    \vspace{-0.5cm}}
    \label{tab:components_in_instruction_data}
\end{table*}

\myparagraph{Analysis of \CaD Instruction Data Components} 
We verify the effectiveness of the components in our instruction data by ablating on the different combinations of our tuning tasks, including: (i) commonality summary (\textit{comm.}); (2) difference summary (\textit{diff.}); and (iii) text-to-image retrieval (\textit{t2i retriev.}) in Table~\ref{tab:components_in_instruction_data}. Training solely on the t2i retrieval task (row B) leads to minimum performance improvement on BISON and SVO Probes, and performance degradation on the three benchmarks of the open-ended tasks due to lacking of any \CaD learning. Training with the commonality (row C) and difference summary (row D) tasks separately lead to a significant boost on the VG comm (2.67) and VG diff (2.89) tasks correspondingly. Training with combinations of the three tasks (F) boosts the performance in comparison to the case of each single component, except for VG comm where the commonality training (row C) leads to better results on this task. Finally, combining phase-1 and phase-2 data (row G) leads to further performance boosts on most of the benchmarks.

\vspace{-0.2cm}
\section{Conclusions, Limitations, and Broader Impact}\label{sec:conclusions}
\vspace{-0.2cm}
We are contributing \OurApproach - an effective, two-phase strategy for collecting \CaDfull (\CaD) Visual Instruction (VI) data, resulting in the also contributed large scale \OurData with 349K samples for verified improvement of \CaD and related image and text comparative capabilities of LMMs. Additionally, we contribute \OurBenchmark - a benchmark of 7.6K open-ended QA to directly evaluate \CaD capabilities between pairs of images. We extensively evaluate and validate our \OurApproach approach, showing it leads to substantial improvements in \CaD abilities and related tasks. We further show how the very few existing \CaD resources are complementary to our approach and can be further refined automatically using our \OurApproach. We believe that our work contributes to the important investigation and improvement of (currently somewhat missing) \CaD abilities of modern LMMs and leads to exciting future work of \CaD VI tuning.\\
\myparagraph{Limitations} 
Currently, our \OurApproach only focuses on the \CaD between two images, and we leave the extension of understanding \CaD and group relations on three or more images to future work.\\
\myparagraph{Broader Impact}
Our \OurApproach, \OurData, and \OurBenchmark significantly contribute to the understanding and improvement of \CaD capabilities in LMMs, and are intended to enhance the applicability and utility of AI across various fields, from robotics to industrial applications. However, this LMM improvement could also lead to job displacement, as these models could increasingly automate complex tasks traditionally performed by humans.

\section{Acknowledgments}
We acknowledge EuroHPC Joint Undertaking for awarding us access to Karolina at IT4Innovations,
Czech Republic, Leonardo at CINECA, Italy, MeluXina at LuxProvide, Luxembourg and LUMI at
CSC, Finland.

\appendix
\section*{Appendix}

In the appendix, we first introduce our dataset release (Sec.~\ref{sec:dataset_release}) and the list of assets (Sec.~\ref{sec:list_assets}) used in this project. 

For further insights into our approach \OurMethod, we report more statistics on our generated data (Sec.~\ref{sec:generated_data_statistics}), and statistics on the external evaluation datasets (Sec.~\ref{sec:statistics_of_external_evaluation_datasets}). We provide more implementation details (Sec.~\ref{sec:implementation_details}) including the specifics of baseline methods, data generation, training and evaluation details. 

As additional results, we report the error bars (Sec.\ref{sec:error_bars}), analyze the Phase-2 data collection on Out-Of-Distribution data (Sec.~\ref{sec:phase2_data_collect_ood_cad_refinement}). Finally, we show qualitative results of the collected \CaD summaries (Sec.~\ref{sec:qualitative_cad_summary}), and compare LMM predictions on our \OurBenchmark benchmark (Sec.~\ref{sec:qualitative_results_on_cad_qa}), and LMM predictions on the BISON dataset (Sec.~\ref{sec:qualitative_results_on_bison}).



\section{Dataset Release}\label{sec:dataset_release}
HuggingFace dataset repository: \url{https://huggingface.co/datasets/wlin21at/CaD-Inst}

We upload both the instruction-following dataset \OurData and the evaluation benchmark \OurBenchmark to the HuggingFace dataset repository at \url{https://huggingface.co/datasets/wlin21at/CaD-Inst} in the repository.

The dataset card with the dataset viewer, dataset description with intended use, and instructions for loading the dataset and downloading the image sources are available.

The dataset repository will be hosted in a long term with data access and necessary maintenance ensured.








\section{List of Assets}\label{sec:list_assets}
Our image sources and annotations are obtained from public datasets. We release our data in accordance to the source data licenses. 

Here is a list of image sources: 
\begin{itemize}
  \item Open Images v6~\cite{kuznetsova2020open} (\url{https://storage.googleapis.com/openimages/web/download_v6.html}): The images are under Creative Commons Attribution (CC BY) 2.0 license. 
  \item COCO 2017~\cite{chen2015microsoft,lin2014microsoft} (\url{https://cocodataset.org/#download}): The images are under a Creative Commons Attribution 4.0 license. 
  \item Flicker30K~\cite{young2014image} (\url{https://shannon.cs.illinois.edu/DenotationGraph/}): The images are the property of SmugMug or its third party licensors and are protected by United States and international intellectual property laws. The images are provided for researchers and educators who wish to use the dataset for non-commercial research and/or educational purposes. 
  \item ADE20K~\cite{zhou2019semantic} (\url{https://groups.csail.mit.edu/vision/datasets/ADE20K/index.html#Download}): The images belong to MIT CSAIL and are licensed under a Creative Common BSD-3 License. 
  \item Visual Genome~\cite{krishna2017visual} (\url{https://homes.cs.washington.edu/~ranjay/visualgenome/api.html}): The images are under a Creative Commons Attribution 4.0 license. 
\end{itemize}
Here is a list of image annotation sources: 
\begin{itemize}
    \item Localized narratives~\cite{pont2020connecting} (\url{https://google.github.io/localized-narratives/}): The annotations are released under a Creative Common Attribution (CC BY) 4.0 license.
    \item MIMIC-IT~\cite{li2023mimic} (\url{https://huggingface.co/datasets/pufanyi/MIMICIT}): The annotations are released under an MIT license. 
    \item SVIT~\cite{zhao2023svit} (\url{https://huggingface.co/datasets/BAAI/SVIT}): The annotations are licensed under a Creative Commons Attribution 4.0 license. It should abide by the policy of OpenAI (\url{https://openai.com/policies/terms-of-use}). The use of original images and annotations from Visual Genome and MS-COCO should comply with the original licenses. 
\end{itemize}
Here is a list of implementation sources or model weights:
\begin{itemize}
    \item LLaVA~\cite{llava,llava1_5} (\url{https://github.com/haotian-liu/LLaVA}): The code is released under an Apache-2.0 license. The project utilizes certain datasets and checkpoints that are subject to their respective original licenses, including but not limited to the OpenAI Terms of Use\footnote{\url{https://openai.com/policies/eu-terms-of-use/}} for the dataset and the specific licenses for base language models for checkpoints trained using the dataset (\eg LLaMA community license\footnote{\url{https://ai.meta.com/llama/license/}} for LLaMA-2 and Vicuna-v1.5). 
    \item Mixtral 8$\times$7B model~\cite{jiang2024mixtral} (\url{https://huggingface.co/mistralai/Mixtral-8x7B-v0.1}): The model is released under an Apache-2.0 license. Usage is subject to the term of use for Mistral products and services\footnote{\url{https://mistral.ai/terms/\#terms-of-use}}.
\end{itemize}

\section{Dataset Statistics}\label{sec:dataset_statistics}



\subsection{Generated Data Statistics}\label{sec:generated_data_statistics}

\begin{figure*}[ht]
\includegraphics[width=\columnwidth]{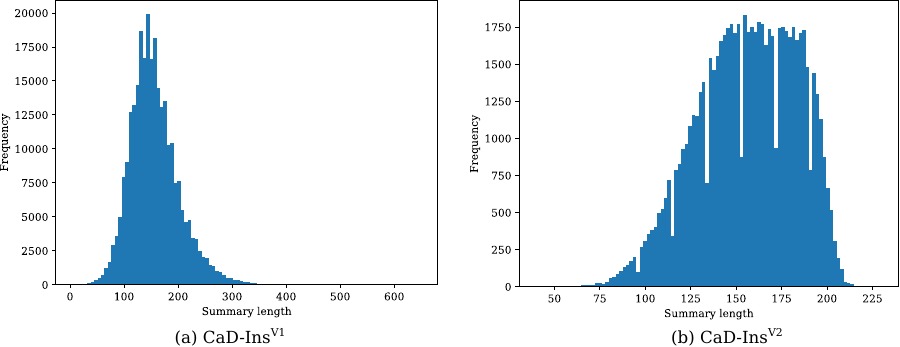}
\caption{
\label{fig:summary_len_distribution}
Distribution of length of \CaD summaries (in terms of number of words) in (a) \OurDataPhaseOne and (b) \OurDataPhaseTwo
}
\end{figure*}

\myparagraph{\OurDataPhaseOne and \OurDataPhaseTwo.}
In \OurDataPhaseOne, we collected structured summaries of \CaD for 278K image pairs, with an average length of 157 words (40 for commonalities and 117 for differences). In \OurDataPhaseTwo, we collected summaries of \CaD for 71K images pairs used in Scene-Difference~\cite{li2023mimic}, with an average length of 156 words (28 for commonalities and 128 for differences). We demonstrate the distribution of CaD summary length (number of words) in \OurDataPhaseOne (Fig.~\ref{fig:summary_len_distribution}(a)) and in \OurDataPhaseTwo (Fig.~\ref{fig:summary_len_distribution}(b)).

\begin{figure*}[ht]
\includegraphics[width=\columnwidth]{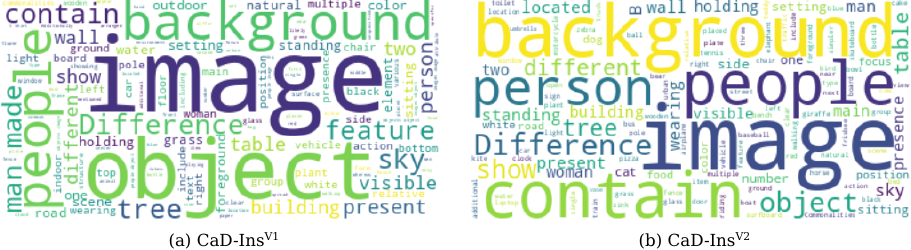}
\caption{
\label{fig:word_cloud}
Word clouds of \CaD summaries in (a) \OurDataPhaseOne and (b) \OurDataPhaseTwo
}
\end{figure*}

In Fig.~\ref{fig:word_cloud}, we also illustrate the cloud of words covered in the \CaD summaries in \OurDataPhaseOne (Fig.~\ref{fig:word_cloud}(a)) and in \OurDataPhaseTwo (Fig.~\ref{fig:word_cloud}(b)). 

\begin{figure*}[ht]
\includegraphics[width=0.7\columnwidth]{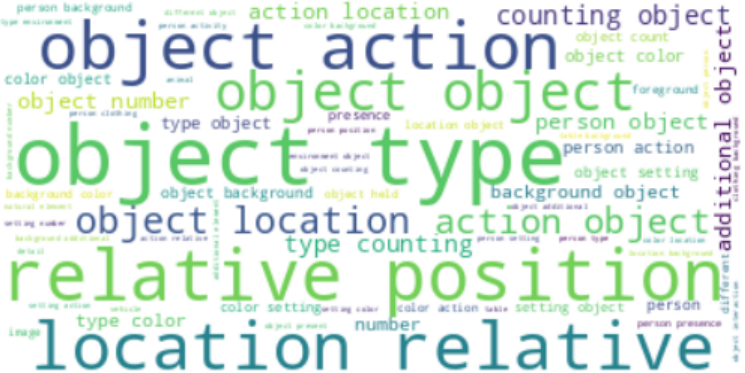}
\caption{
\label{fig:word_cloud_cad_characteristics}
Word cloud of sample-specific characteristics in \CaD summaries in \OurDataPhaseOne. The distribution of these sample-specific characteristics is also shown in a Sunburst chart in Fig.~3(a)(main paper).
}
\end{figure*}
In the main paper, we mentioned that the collected summaries are structured according to approximate 6 axes of characteristics: \textit{object types, attributes, counting, actions, locations} and \textit{relative positions}. Note that the characteristics appear unevenly on a case-to-case basis based on the LLM decision on individual samples. In Fig.~3(a)(main paper), we illustrate the distribution of these sample-specific characteristics in a Sunburst chart. Here in Fig.~\ref{fig:word_cloud_cad_characteristics}, we also illustrate the cloud of words in these characteristics in \CaD summaries in the Phase-1 data collection \OurDataPhaseOne.

\begin{figure*}[ht]
\includegraphics[width=\columnwidth]{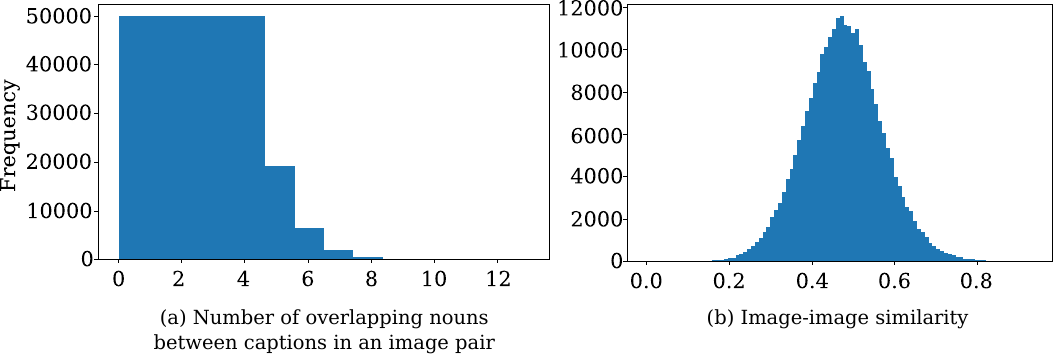}
\caption{
\label{fig:phase1_img_img_sim_n_overlap_noun}
Distribution of (a) number of overlapping nouns between captions in an image pair and (b) image-image similarities in the 278K image pairs in \OurDataPhaseOne
}
\end{figure*}

In the main paper, we introduced that we collect 278K image pairs with different levels of similarity between their captions. We measure the similarity between two captions by counting the number of overlapping nouns in the corresponding captions. 
Here we show the distribution of the number of overlapping nouns in Fig.~\ref{fig:phase1_img_img_sim_n_overlap_noun}(a). We see that we cover image pairs with different levels of caption-caption similarity. Furthermore, we use the CLIP ViT-B/32 model~\cite{clip} to compute the similarity scores between the two images in each pair and report the distribution in Fig.~\ref{fig:phase1_img_img_sim_n_overlap_noun}(b). We verify that image pairs of diverse similarity levels are covered in our Phase-1 data collection \OurDataPhaseOne.




\begin{figure*}[ht]
\includegraphics[width=\columnwidth]{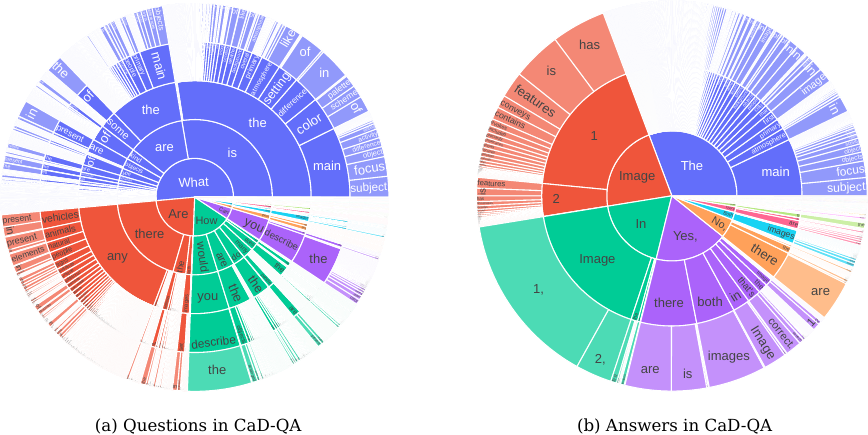}
\caption{
\label{fig:CaD_QA_distribution}
 Distribution of (a) questions (first 5 words) and (b) answers (first 3 words) in the evaluation benchmark \OurBenchmark.
}

\end{figure*}

\myparagraph{\OurBenchmark.} Our \OurBenchmark benchmark contains 7.5K open-ended questions with answers. Here we show the distribution of questions types (first 5 words) and answer types (first 3 words) in Sunburst charts in Fig.~\ref{fig:CaD_QA_distribution}. There are diverse question categories covered such as \textit{Yes/No} questions, \textit{What} questions on scene characteristics such as objects, attributes and setting, and also requests to describe specific characteristics in details. 




\subsection{Statistics of External Evaluation Datasets}\label{sec:statistics_of_external_evaluation_datasets}
We evaluate on several external VQA benchmarks of closed-ended and open-ended questions. Here we give a brief introduction on the contents and statistics. 

\myparagraph{BISON} is a dataset for the binary image selection task~\cite{hu2019evaluating}. There are 150 samples in the evaluation benchmark, each sample consisting of a pair of two visually similar images and a query caption. Only one image correctly matches with the query caption. It measures the ability of the LMMs to relate fine-grained text content in the caption to visual content in the images. 

\myparagraph{SVO Probes} is a benchmark designed to probe for subject, verb and object understanding in vision-language models~\cite{hendricks2021probing}. In the benchmark, each sample consists of a pair of two images and a query sentence, where only one image correctly matches with the query sentence. The negative image differs from the positive image with regard to either the subject, the verb or the object. There are 36.8K samples in the dataset. For efficient evaluation, we randomly select 1500 samples that can be divided into 3 partitions \textit{subject}, \textit{verb} and \textit{object} where each partition has 500 samples with the image pair contradiction in either subject, verb or object. 

\myparagraph{EQBEN} is a benchmark that focuses on visual minimal change between two images~\cite{wang2023equivariant}. Each sample in the benchmark consists of a pair of two images with subtle visual changes and two corresponding captions. The dataset is comprised of frames from natural video datasets such as YouCook2~\cite{zhou2018towards}, Action Genome~\cite{ji2020action} and GEBC~\cite{wang2022geb+}, as well as sythetic image pairs with subtle differences generated by the photo-realistic scene generator Kubric~\cite{greff2022kubric} and the diffusion model Stable-Diffusion~\cite{rombach2022high}. We employ an EQBEN subset\footnote{\url{https://entuedu-my.sharepoint.com/:u:/g/personal/tan317_e_ntu_edu_sg/ETkpKSsmun1MpBw7FqfUUS8BwTX2gKkTQkDFsfOGCw-9yA?e=KGtpg0}} which is released by the authors in~\cite{wang2023equivariant} for evaluating the performance of LMMs specifically. The subset consists of 120 samples, comprised of frame pairs from Action Genome and GEBC, image pairs with changes in attributes, count and location generated by Kubric, and image pairs with style change generated by Stable-Diffusion. For each sample, we perform the binary image selection task twice, feeding one of the descriptions for image selection at a time. The sample is considered positively answered only when both selection tasks are correctly solved.

\myparagraph{COLA} is a benchmark for evaluating the capabilities of vision-language models on representing simple compositions by combing objects with their attributes~\cite{ray2023cola}. Each sample in the benchmark consists of two images with two corresponding captions. The two images have attributes and objects that are swapped in the captions, \eg \textit{large tree to the right of little short green tree},  and \textit{tall green tree to the right of large tall green tree}. We employ the partition of \textit{multi-object setting} in the benchmark which consists of 210 image pairs and captions. Similar to evaluation on EQBEN, we perform the binary image selection task twice for each sample. 

\myparagraph{NLVR2} is a benchmark for evaluation of the visual reasoning with natural language task which aesses the ability of LMMs to predict whether a sentence is true about a pair of images~\cite{suhr2019corpus}. The task focuses on understanding of compositionalities in terms of relations, comparisons and counting. We use the subset of 150 samples provided in SparklesChat~\cite{huang2023sparkles} for a fair comparison.

\myparagraph{SEED-Bench} is an evaluation benchmark on comprehensive vision-language understanding, consisting of 19K multiple choice questions~\cite{li2023seed}. The are two major categories in the benchmark: \textit{SEED-Image} with 14K samples and \textit{SEED-Video} with 5K samples. SEED-Image consists of 9 dimensions: scene understanding, instance identity, instance attributes, instance location, instance counting, spatial relation, visual reasoning and text understanding. All samples contain only a single input image. SEED-Video consists of 3 dimensions: action recognition, action prediction and procedure understanding. The videos are from Something-Something-v2~\cite{goyal2017something}, EPIC-Kitchen~\cite{damen2022rescaling} and Breakfast~\cite{kuehne2014language}.


\section{Implementation Details}\label{sec:implementation_details}

\subsection{Baselines}

\myparagraph{SparklesChat~\cite{huang2023sparkles}} is finetuned from the first-stage pretrained model of MiniGPT4~\cite{minigpt4}. The model is finetuned with their collected multi-image dialogue data. SparklesChat follows the architecture of MiniGPT4 and uses Vicuna 7B~\cite{vicuna}, EVA-CLIP ViT-G/14~\cite{fang2023eva} with a Q-Former from BLIP-2~\cite{li2023blip}. We use the model weights and instruction templates available at \url{https://github.com/HYPJUDY/Sparkles}. 

\myparagraph{Otter~\cite{li2023otter}} is finetuned from the OpenFlamingo model~\cite{awadalla2023openflamingo} with the collected multimodal in-context instruction-response data in MIMIC-IT~\cite{li2023mimic}. We use their most recent open-sourced version Otter-Image-LLaMA7B-LA-InContext available at \url{https://huggingface.co/luodian/OTTER-Image-LLaMA7B-LA-InContext}.

\myparagraph{MMICL~\cite{zhao2024mmicl}} is based on the InstructBLIP model~\cite{instructblip}. The model is finetuned their own collected multimodal in-context learning datast consisting of interleaved text-image inputs, inter-related multiple image inputs and multimodal in-context learning inputs. We evaluate with their model of the largest scale MMICL-InstructBLIP-T5-XXL, available at \url{https://huggingface.co/BleachNick/MMICL-Instructblip-T5-xxl}. 

\myparagraph{EMU2-Chat~\cite{sun2023generative}} is a generative multimodal model trained on large-scale multimodal sequences. The model consists of pretrained  EVA-02-CLIP-E-plus~\cite{sun2023eva} and LLaMA-33B~\cite{llama}. The model weights and inference code are available at~\url{https://huggingface.co/BAAI/Emu2-Chat}. 

\myparagraph{InternLM-XComposer2-VL~\cite{zhang2023internlm}} consists of CLIP ViT-L~\cite{clip} and InternLM2-7B~\cite{team2023internlm}. The model weights of the InternLM-XComposer2-VL-7B and inference code are available at~\url{https://huggingface.co/internlm/internlm-xcomposer2-vl-7b}.

\myparagraph{LLaVA 1.5~\cite{llava1_5}} is an improved version from LLaVA~\cite{llava} with CLIP-ViT-L-336px~\cite{clip} as the visual backbone and Vicuna 1.5~\cite{zheng2023judging} as the LLM. Our visual instruction tuning is performed using the open-sourced code of LLaVA 1.5. We train on the first-stage pretrained weights of LLaVA 1.5 via LoRA finetuning. We evaluate both LLaVA 1.5 7B lora and LLaVA 1.5 13B lora as baselines. The models are available at \url{https://huggingface.co/liuhaotian/llava-v1.5-7b-lora} and \url{https://huggingface.co/liuhaotian/llava-v1.5-13b-lora}. 

\myparagraph{LLaVA 1.6~\cite{liu2024llavanext}} is an improved version from LLaVA 1.5 with increased input image resolution and improved mixture of instruction tuning data. The 7B and 13B versions are avaible on Huggingface at \url{https://huggingface.co/liuhaotian/llava-v1.6-vicuna-7b} and \url{https://huggingface.co/liuhaotian/llava-v1.6-vicuna-13b}. However, the training code is not yet available.

\subsection{Implementation Details}

\begin{figure*}[ht]
\includegraphics[width=\columnwidth]{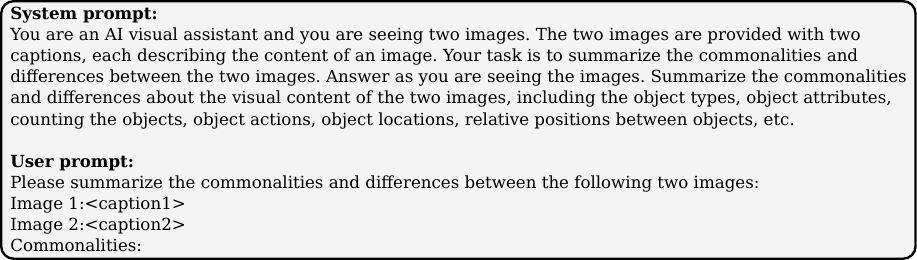}
\caption{
\label{fig:prompt_for_cad_summary}
Prompt for the task of Phase-1 LLM-based \CaD summary. 
}
\end{figure*}

\myparagraph{Data Collection.} In Phase-1, we leverage the Mixtral 8x7B Instruct v0.1 model\footnote{Huggingface source: \url{https://huggingface.co/mistralai/Mixtral-8x7B-Instruct-v0.1}} with 8-bit inference for data generation. We set the batch size to 16 and max new token to 750. The prompt for the task of LLM-based \CaD summary is given in Fig.~\ref{fig:prompt_for_cad_summary}. The generation with batch 16 fits to an A100 80G GPU.

\begin{figure*}[ht]
\includegraphics[width=\columnwidth]{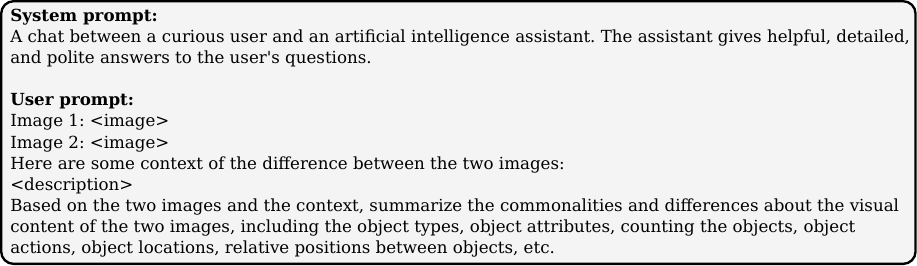}
\caption{
\label{fig:prompt_for_cad_summary_phase2}
Prompt for the task of Phase-2 LMM-based \CaD summary. 
}
\end{figure*}

In Phase-2, we leverage the Phase-1 model \OurModelPhaseOne 13B model to generate \CaD summary on additional image pairs. The temporature, max new tokens and number of beams are set to 0, 256 and 1. The prompt for the task of LMM-based \CaD summary is given in Fig.~\ref{fig:prompt_for_cad_summary_phase2}.

\begin{figure*}[!ht]
\includegraphics[width=\columnwidth]{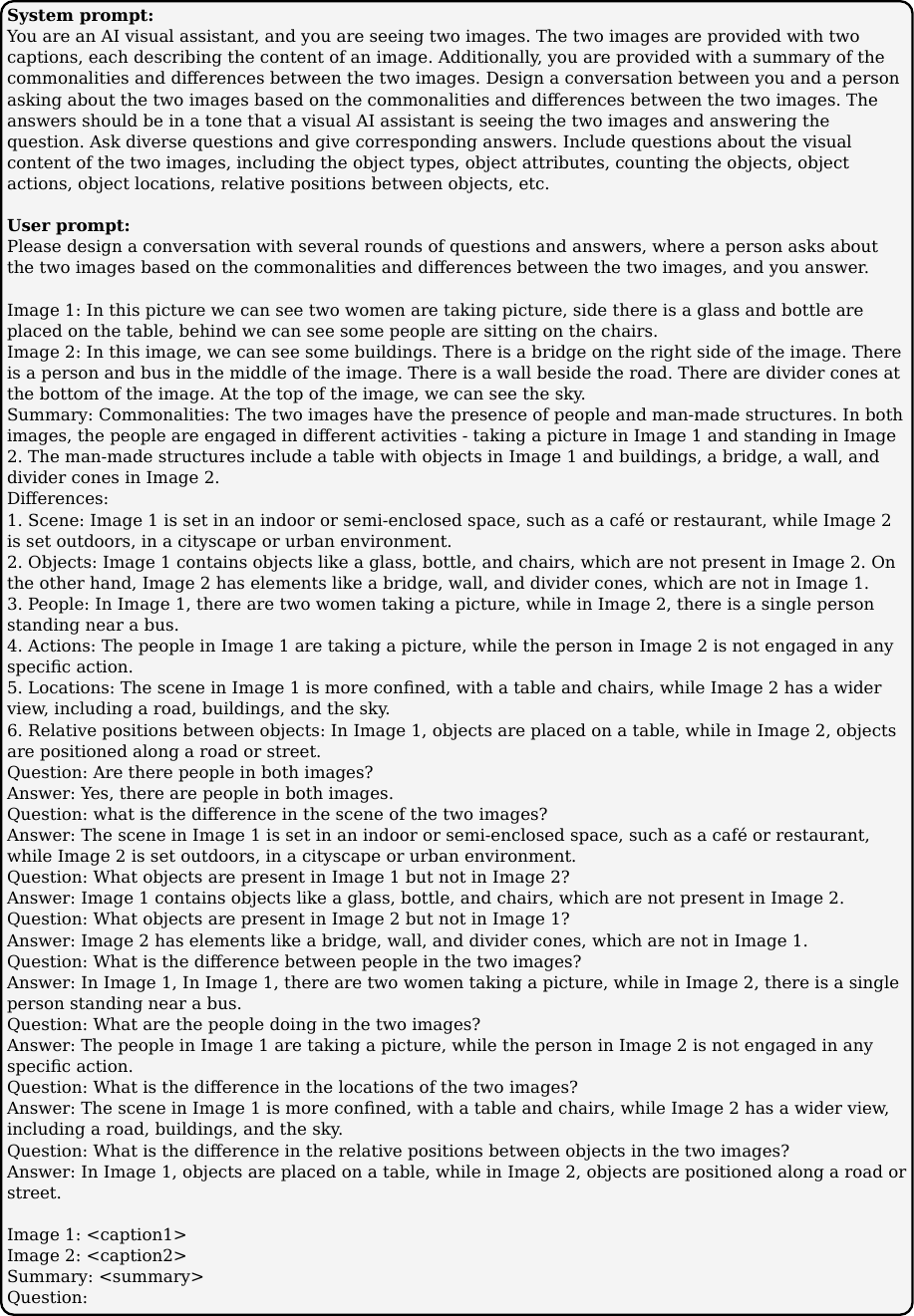}
\caption{
\label{fig:prompt_for_cad_qa_generation}
Prompt for the task of generating Q\&A pairs based on both image captions and the \CaD summary.
}
\end{figure*}

For collecting open-ended QAs in \OurBenchmark, we first use the LMM to generate the \CaD summaries based on the image captions (see Fig.~\ref{fig:prompt_for_cad_summary}). Then we prompt the LLM with both the image captions and the \CaD summary, instructing it to generate a multi-turn conversation with several rounds of Q\&A. We also provide some in-context samples to demonstrate the desired layout. The prompt for the task of generating Q\&A pairs based on both image captions and the \CaD summary is illustrated in Fig.~\ref{fig:prompt_for_cad_qa_generation}.


\myparagraph{Training.} We perform visual instruction tuning following the configuration in LLaVA 1.5. We set the batch size to 128 and train for one epoch. The learning rate for LLM with LoRA and for the projector are set to $1\times 10^{-4}$ and $2\times 10^{-5}$ correspondingly. The LoRA rank and alpha values are set to 128 and 256. The training experiments are run on 4$\times$A100 80G GPUs.

\myparagraph{Inference.} For VQA inference, the temperature, max new tokens and number of beams are set to 0, 256 and 1.

\begin{figure*}[ht]
\includegraphics[width=\columnwidth]{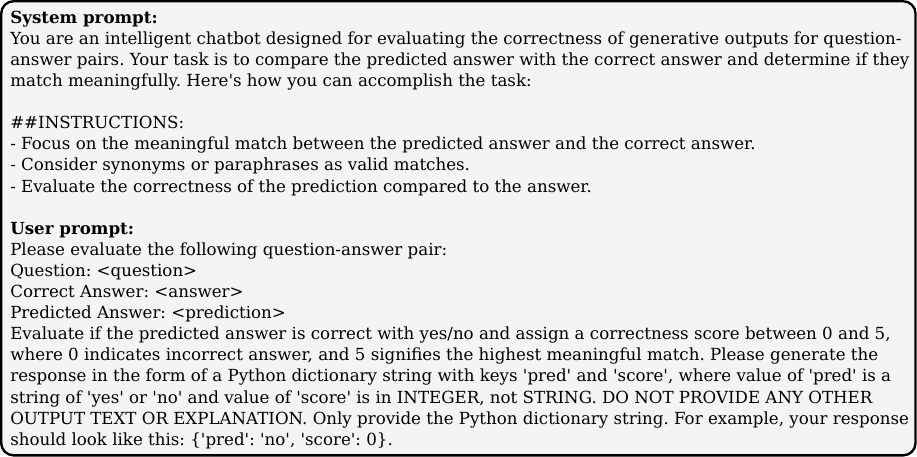}
\caption{
\label{fig:prompt_for_llm_evaluation}
Prompt for the LLM-assisted evaluation. 
}
\end{figure*}

\myparagraph{LLM-assisted Evaluation} We leverage the Mixtral 8$\times$7B model for LLM-assisted evaluation on open-ended questions. We feed the question, correct answer and the predicted answer into the LLM and instruct it to provide a rating between 0 and 5. The prompt for generating the evaluation rating is given in Fig.~\ref{fig:prompt_for_llm_evaluation}.

\section{Additional Results}\label{sec:additional_results}





\subsection{Error Bars}\label{sec:error_bars}
\begin{table*} [!tb]
    \centering
    \scriptsize
    \begin{tabular}{cccccccccccc}
    \toprule
    Training Data  &  BISON &  SVO  &  EQBEN & COLA & \OurBenchmark \\
    \midrule
    LLaVA mix + \OurModelPhaseOne &  91.78\% $\pm$ 1.02\% & 92.33\% $\pm$ 0.57\% & 33.06\% $\pm$ 0.96\% & 34.64\% $\pm$ 2.09\% & 3.270 $\pm$ 0.002  \\

    \bottomrule
    \end{tabular}
    \caption{ 
    Average performance of the Phase-1 model \OurModelPhaseOne on multiple runs of training. 
     }
    \label{tab:error_bar}
\end{table*}
We run the training of the Phase-1 model \OurModelPhaseOne multiple times and report the average performance with standard deviation in Table~\ref{tab:error_bar}. In most evaluation cases, the standard deviation is within around 1\%.

\subsection{Ablation on Phase-2 Data Collection - OOD CaD refinement}\label{sec:phase2_data_collect_ood_cad_refinement}
\begin{table*} [!tb]
    \centering
    \scriptsize
    \begin{tabular}{M{0.05cm}cM{0.9cm}M{0.9cm}M{1.2cm}M{1.3cm}}
    \toprule
    & Training Data  &  BISON &  SVO  &  Difference Spotting & \OurBenchmark \\
    \midrule

   A:& LLaVA mix (L)    & \textbf{54.00\%} & 46.80\% &   49.50\%  & \underline{2.54}\\
   B:& L + SpotDiff orig. annot.  & \underline{51.33\%}  & \underline{52.27\%} &  \underline{60.48\%} & 2.51   \\
   C:& L + SpotDiff our annot. (refined from orig. annot.) & \textbf{54.00\%} & \textbf{54.87\%} &  \textbf{66.67\%} & \textbf{2.86} \\

    \bottomrule
    \end{tabular}
    \caption{ 
    Ablation of phase-2 data collection from 15K pairs of video frames in Spot-the-diff (SpotDiff). We use \OurModelPhaseOne to generate CaD on SpotDiff by refining from the original human-annotated difference descriptions. 
    }
    \label{tab:train_on_spot_the_diff}
\end{table*}
In Section 6 (main paper), we perform ablation the Phase-2 data collection. Here we further explore applying our phase-2 data collection on out-of-distribution (OOD) data of Spot-the-diff (SpotDiff) dataset. The dataset contains distant-view frame pairs with very subtle changes from video-surveillance footage, which are OOD from most LMM training data. 

In Table~\ref{tab:train_on_spot_the_diff}, we train with SpotDiff original human-annotated difference description (row B) and with our \OurModelPhaseOne generated \CaD summaries which is refined from the original annotation (row C). 
We also evaluate on the Difference-Spotting partition on SEED-Bench 2~\cite{li2023seed2} which contains multi-choice questions based on frame pairs from SpotDiff. In data collection and training for this experiment, we only used the 15K training image pairs from SpotDiff which are not included in the Difference-Spotting SEED partition. The results in Table~\ref{tab:train_on_spot_the_diff} verify that our phase-2 data collection using \OurModelPhaseOne is also effective on OOD data. 

\subsection{Qualitative Results of \CaD Summaries}\label{sec:qualitative_cad_summary}

\begin{figure*}[!ht]
\includegraphics[width=\columnwidth]{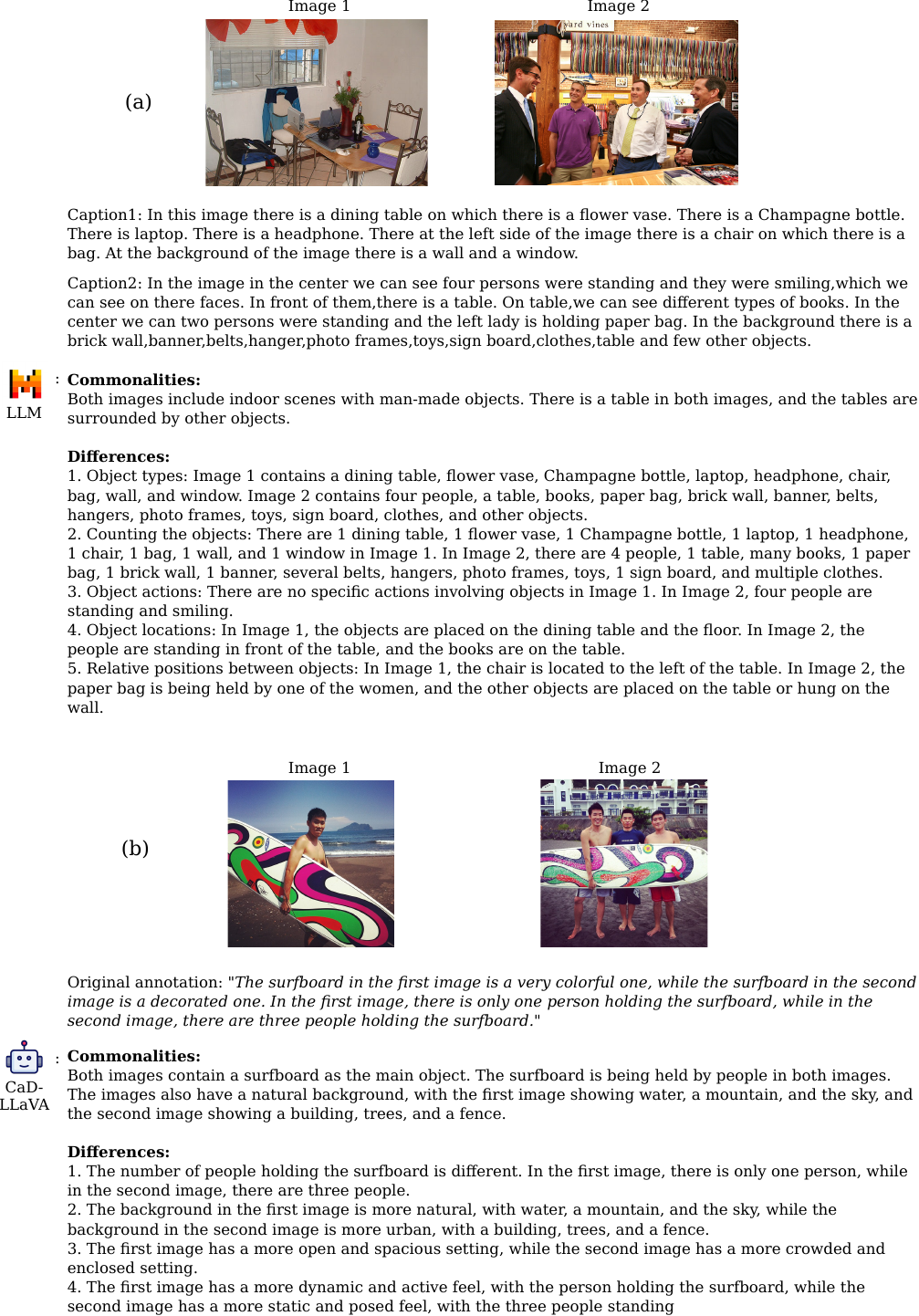}
\caption{
\label{fig:phase1_phase2_data_collection}
Examples of (a) Phase-1 LLM-collected \CaD summary and (b) Phase-2 LMM-collected \CaD summary 
}
\end{figure*}

In Fig.~2 (main paper), we illustrate the pipeline of our two-phase \OurMethod together with two examples of Phase-1 LLM-collected \CaD summary and Phase-2 LMM-collected \CaD summary. Here in Fig.~\ref{fig:phase1_phase2_data_collection}, we provide two additional examples. Note that in Fig.~\ref{fig:phase1_phase2_data_collection}(a), we only pass the captions with the instruction prompt (in Fig.~\ref{fig:prompt_for_cad_summary}) into the LLM. In Fig.~\ref{fig:phase1_phase2_data_collection}(b), we pass the original annotation and both images with the instruction prompt (in Fig.~\ref{fig:prompt_for_cad_summary_phase2}) into the Phase-1 model. In the main paper (Table 5), we demonstrate the generated \CaD summary without using the original annotation also leads to effective results.

\begin{figure*}[!ht]
\includegraphics[width=\columnwidth,height=22cm]{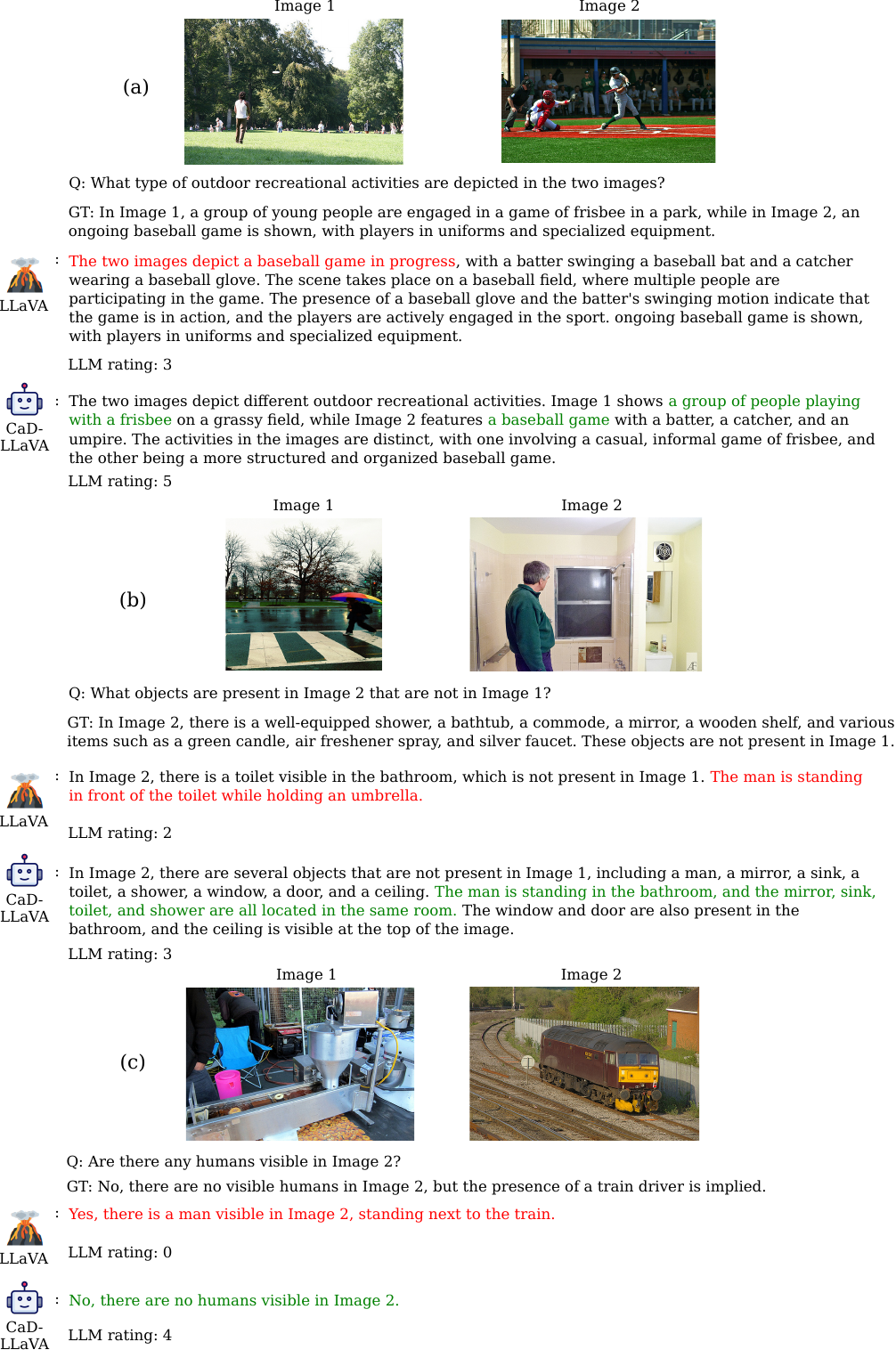}
\caption{
Examples of Q\&A pairs in \OurBenchmark together with LMM predicted answers and the corresponding LLM evaluation rating for the prediction (Red and green texts denote incorrect and correct description). 
}\label{fig:cad_qa_visual_part1}
\end{figure*}

\begin{figure*}[!ht]
\includegraphics[width=\columnwidth,height=22cm]{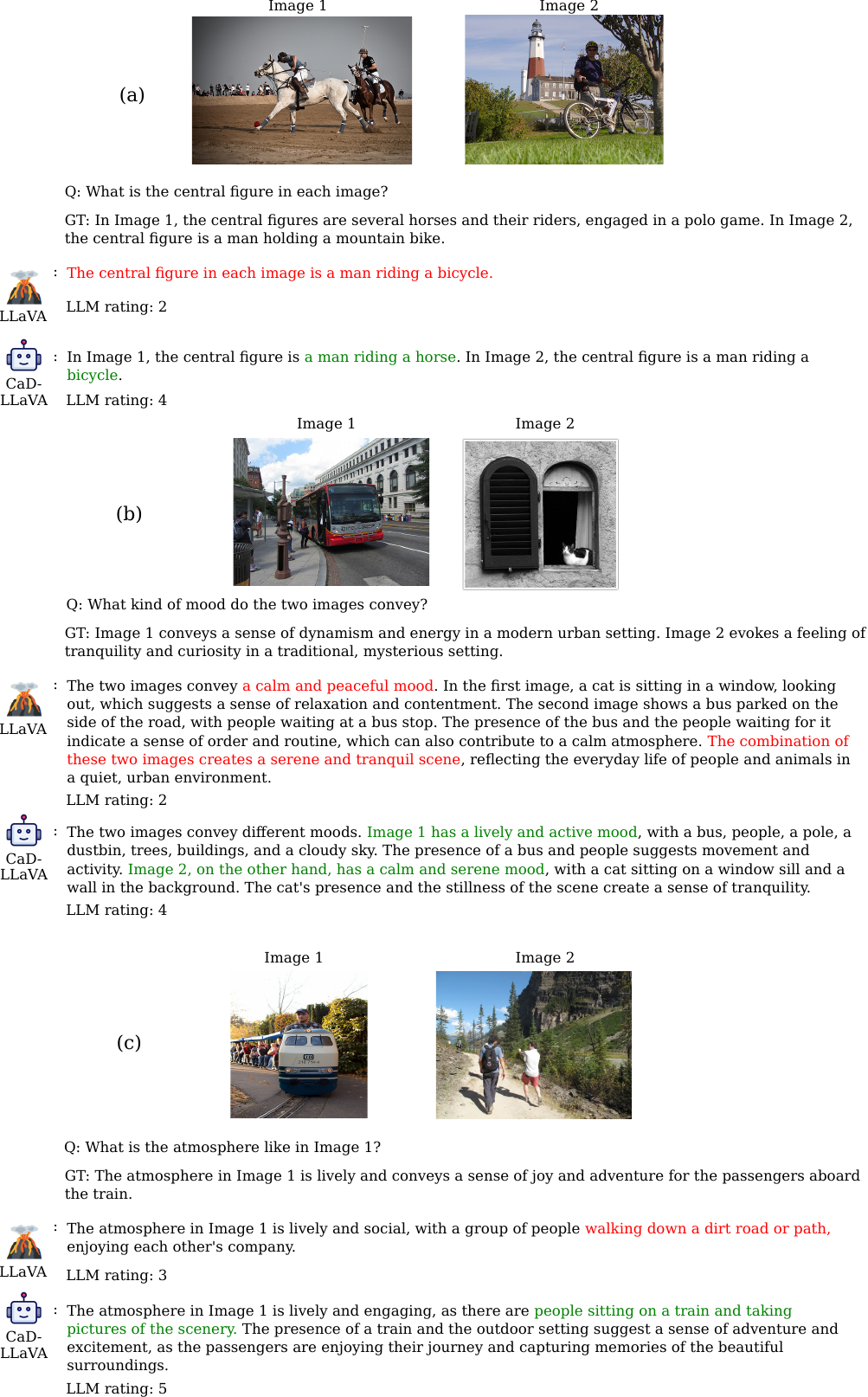}
\caption{
Examples of Q\&A pairs in \OurBenchmark together with LMM predicted answers and the corresponding LLM evaluation rating for the prediction (Red and green texts denote incorrect and correct description). 
}\label{fig:cad_qa_visual_part2}
\end{figure*}

\begin{figure*}[!ht]
\includegraphics[width=\columnwidth,height=22cm]{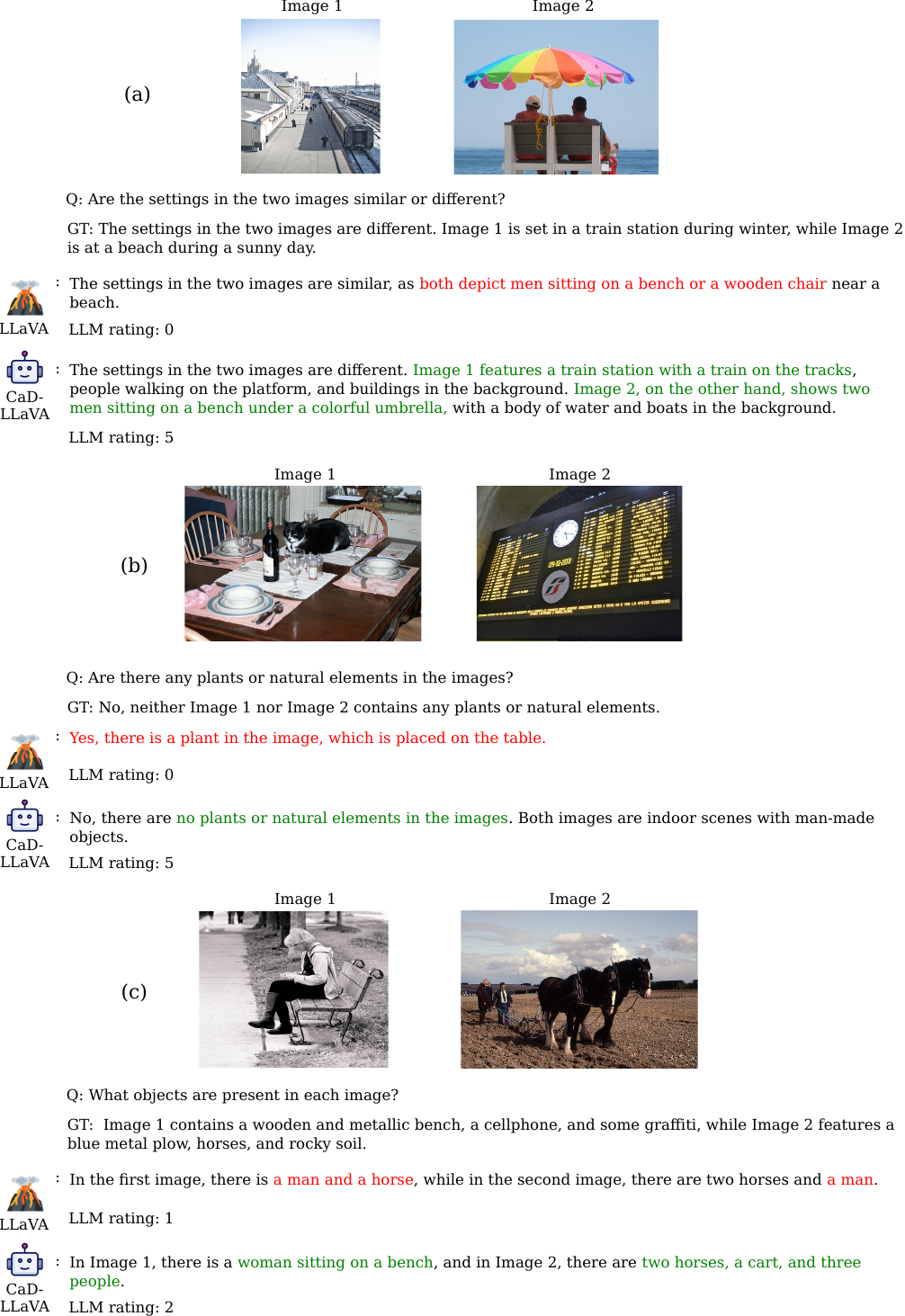}
\caption{
Examples of Q\&A pairs in \OurBenchmark together with LMM predicted answers and the corresponding LLM evaluation rating for the prediction (Red and green texts denote incorrect and correct description). 
}\label{fig:cad_qa_visual_part3}
\end{figure*}

\subsection{Qualitative Results on \OurBenchmark}\label{sec:qualitative_results_on_cad_qa}
In Fig.~\ref{fig:cad_qa_visual_part1}, Fig.~\ref{fig:cad_qa_visual_part2} and Fig.~\ref{fig:cad_qa_visual_part3}, we show examples of Q\&A pairs in our \OurBenchmark, together with the predicted answers from \OurModelPhaseTwo model and the vanilla LLaVA 1.5 model. We also report the LLM ratings for the predicted answers. The vanilla LLaVA model has incorrect answers by either mistakenly combining the contents in two images (Fig.~\ref{fig:cad_qa_visual_part1}(b), \textit{the man is standing in front of the toilet while holding an umbrella}), omitting one of the images (Fig.~\ref{fig:cad_qa_visual_part2}(a), Fig.~\ref{fig:cad_qa_visual_part3}(a)), attending to the incorrect image (Fig.~\ref{fig:cad_qa_visual_part2}(c)) or hallucinating non-existent contents (Fig.~\ref{fig:cad_qa_visual_part3}(b)). The failure demonstrates the lacking of capability of properly comparing two images. At the same time, our \OurDataPhaseTwo manages to correctly differentiate between the two images, attend to the corresponding content asked in the question and draw a summary of comparison.  




\subsection{Qualitative Results on BISON}\label{sec:qualitative_results_on_bison}
\begin{figure*}[!ht]
\includegraphics[width=\columnwidth]{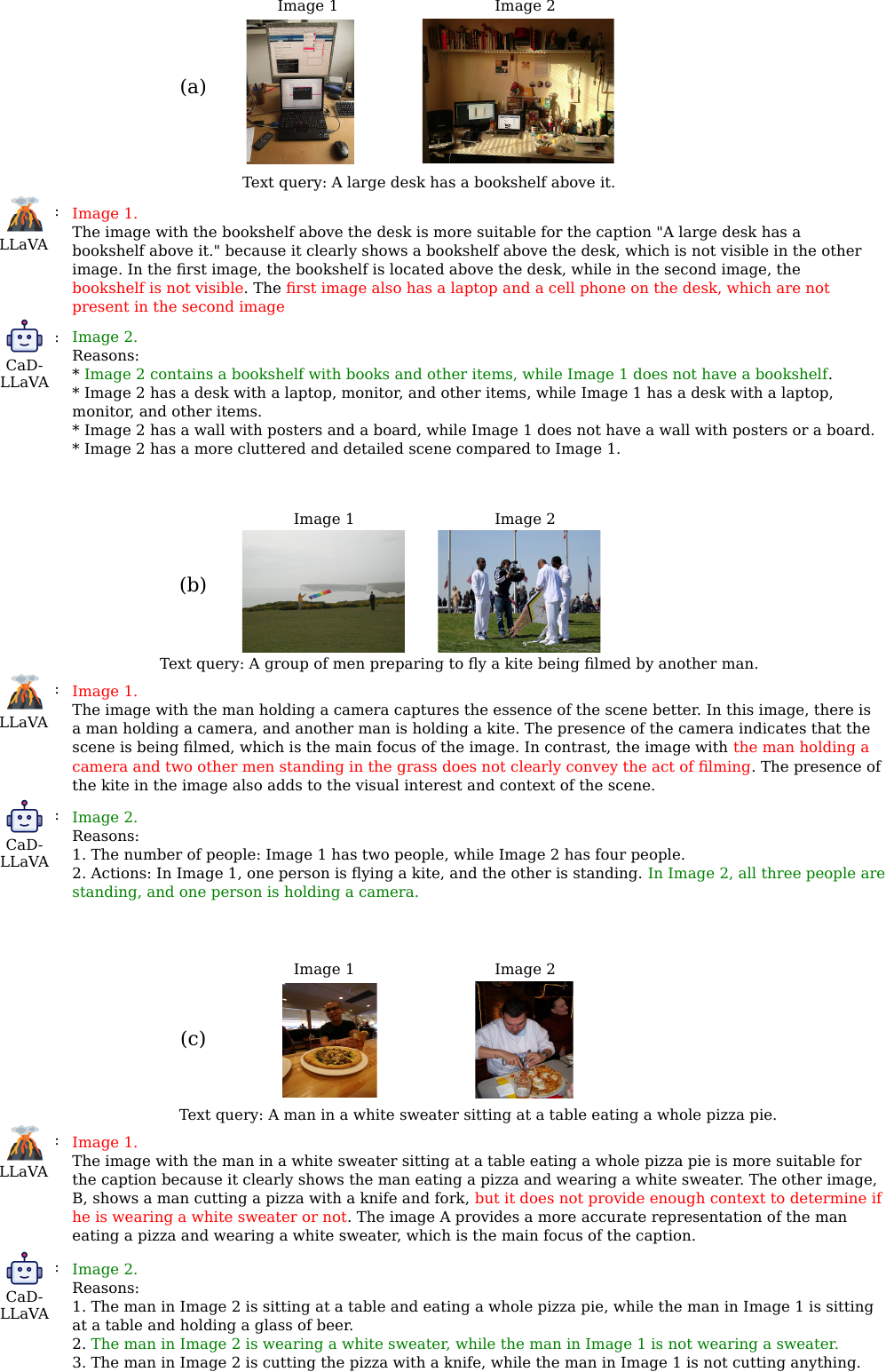}
\caption{
Examples of predictions of the binary image selection task on BISON (red and green texts denote incorrect and correct predictions). We instruct the LMMs to, besides the selection, also give a reasoning for the answer.
}\label{fig:bison_visualization}
\end{figure*}

In Fig.~\ref{fig:bison_visualization}, we illustrate some examples of the binary image selection task on BISON. We instruct the LMMs to give both the selection answer and also the reasoning for the selection. Here we compare the vanilla LLaVA 1.5 and our \OurModelPhaseTwo. 
The LLaVA model, even if it captures the relevant content in some cases, has confusion differentiating the two images (Fig.~\ref{fig:bison_visualization}(a)(b)). For our \OurModelPhaseTwo, the key reasoning that leads to the correct answer is always covered in the structured difference summary.



\clearpage

{\small
\printbibliography

@String(CVPR= {IEEE Conf. Comput. Vis. Pattern Recog.})

@String(AAAI = {Proc. AAAI})

@String(CVPR  = {Proc. CVPR})

@String(IEEE = {Proc. IEEE})

@inproceedings{clip,
  title={Learning transferable visual models from natural language supervision},
  author={Radford, Alec and Kim, Jong Wook and Hallacy, Chris and Ramesh, Aditya and Goh, Gabriel and Agarwal, Sandhini and Sastry, Girish and Askell, Amanda and Mishkin, Pamela and Clark, Jack and others},
  booktitle={International conference on machine learning},
  pages={8748--8763},
  year={2021},
  organization={PMLR}
}

@article{llama2,
  title={Llama 2: Open foundation and fine-tuned chat models},
  author={Touvron, Hugo and Martin, Louis and Stone, Kevin and Albert, Peter and Almahairi, Amjad and Babaei, Yasmine and Bashlykov, Nikolay and Batra, Soumya and Bhargava, Prajjwal and Bhosale, Shruti and others},
  journal={arXiv preprint arXiv:2307.09288},
  year={2023}
}

@inproceedings{li2023blip,
  title={Blip-2: Bootstrapping language-image pre-training with frozen image encoders and large language models},
  author={Li, Junnan and Li, Dongxu and Savarese, Silvio and Hoi, Steven},
  booktitle={International conference on machine learning},
  pages={19730--19742},
  year={2023},
  organization={PMLR}
}

@article{li2023m,
  title={M3IT: A Large-Scale Dataset towards Multi-Modal Multilingual Instruction Tuning},
  author={Li, Lei and Yin, Yuwei and Li, Shicheng and Chen, Liang and Wang, Peiyi and Ren, Shuhuai and Li, Mukai and Yang, Yazheng and Xu, Jingjing and Sun, Xu},
  journal={arXiv preprint arXiv:2306.04387},
  year={2023}
}

@misc{liu2024llavanext,
    title={LLaVA-NeXT: Improved reasoning, OCR, and world knowledge},
    url={https://llava-vl.github.io/blog/2024-01-30-llava-next/},
    author={Liu, Haotian and Li, Chunyuan and Li, Yuheng and Li, Bo and Zhang, Yuanhan and Shen, Sheng and Lee, Yong Jae},
    month={January},
    year={2024}
}

@misc{gestalt,
    title={What are the Gestalt Principles?},
    url={https://www.interaction-design.org/literature/topics/gestalt-principles},
    author={Interaction Design Foundation - IxDF},
    month={August},
    year={2016}
}

@article{llava1_5,
  title={Improved baselines with visual instruction tuning},
  author={Liu, Haotian and Li, Chunyuan and Li, Yuheng and Lee, Yong Jae},
  journal={arXiv preprint arXiv:2310.03744},
  year={2023}
}

@article{llava, 
  title={Visual instruction tuning},
  author={Liu, Haotian and Li, Chunyuan and Wu, Qingyang and Lee, Yong Jae},
  journal={Advances in neural information processing systems},
  volume={36},
  year={2023}
}

@inproceedings{minigpt4,
  title={MiniGPT-4: Enhancing Vision-Language Understanding with Advanced Large Language Models},
  author={Zhu, Deyao and Chen, Jun and Shen, Xiaoqian and Li, Xiang and Elhoseiny, Mohamed},
  booktitle={The Twelfth International Conference on Learning Representations},
  year={2023}
}

@article{chen2015microsoft,
  title={Microsoft coco captions: Data collection and evaluation server},
  author={Chen, Xinlei and Fang, Hao and Lin, Tsung-Yi and Vedantam, Ramakrishna and Gupta, Saurabh and Doll{\'a}r, Piotr and Zitnick, C Lawrence},
  journal={arXiv preprint arXiv:1504.00325},
  year={2015}
}

@inproceedings{spotthediff,
  title={Learning to Describe Differences Between Pairs of Similar Images},
  author={Jhamtani, Harsh and Berg-Kirkpatrick, Taylor},
  booktitle={Proceedings of the 2018 Conference on Empirical Methods in Natural Language Processing},
  pages={4024--4034},
  year={2018}
}

@article{instructblip,
  title={Instructblip: Towards general-purpose vision-language models with instruction tuning},
  author={Dai, Wenliang and Li, Junnan and Li, Dongxu and Tiong, Anthony Meng Huat and Zhao, Junqi and Wang, Weisheng and Li, Boyang and Fung, Pascale N and Hoi, Steven},
  journal={Advances in Neural Information Processing Systems},
  volume={36},
  year={2023}
}

@article{vicuna,
  title={Vicuna: An open-source chatbot impressing gpt-4 with 90\%* chatgpt quality},
  author={Chiang, Wei-Lin and Li, Zhuohan and Lin, Zi and Sheng, Ying and Wu, Zhanghao and Zhang, Hao and Zheng, Lianmin and Zhuang, Siyuan and Zhuang, Yonghao and Gonzalez, Joseph E and others},
  journal={See https://vicuna. lmsys. org (accessed 14 April 2023)},
  volume={2},
  number={3},
  pages={6},
  year={2023}
}

@article{llama,
  title={Llama: Open and efficient foundation language models},
  author={Touvron, Hugo and Lavril, Thibaut and Izacard, Gautier and Martinet, Xavier and Lachaux, Marie-Anne and Lacroix, Timoth{\'e}e and Rozi{\`e}re, Baptiste and Goyal, Naman and Hambro, Eric and Azhar, Faisal and others},
  journal={arXiv preprint arXiv:2302.13971},
  year={2023}
}

@article{awadalla2023openflamingo,
  title={Openflamingo: An open-source framework for training large autoregressive vision-language models},
  author={Awadalla, Anas and Gao, Irena and Gardner, Josh and Hessel, Jack and Hanafy, Yusuf and Zhu, Wanrong and Marathe, Kalyani and Bitton, Yonatan and Gadre, Samir and Sagawa, Shiori and others},
  journal={arXiv preprint arXiv:2308.01390},
  year={2023}
}

@article{damen2022rescaling,
  title={Rescaling egocentric vision: Collection, pipeline and challenges for epic-kitchens-100},
  author={Damen, Dima and Doughty, Hazel and Farinella, Giovanni Maria and Furnari, Antonino and Kazakos, Evangelos and Ma, Jian and Moltisanti, Davide and Munro, Jonathan and Perrett, Toby and Price, Will and others},
  journal={International Journal of Computer Vision},
  pages={1--23},
  year={2022},
  publisher={Springer}
}

@inproceedings{fang2023eva,
  title={Eva: Exploring the limits of masked visual representation learning at scale},
  author={Fang, Yuxin and Wang, Wen and Xie, Binhui and Sun, Quan and Wu, Ledell and Wang, Xinggang and Huang, Tiejun and Wang, Xinlong and Cao, Yue},
  booktitle={Proceedings of the IEEE/CVF Conference on Computer Vision and Pattern Recognition},
  pages={19358--19369},
  year={2023}
}

@article{gong2023multimodal,
  title={Multimodal-gpt: A vision and language model for dialogue with humans},
  author={Gong, Tao and Lyu, Chengqi and Zhang, Shilong and Wang, Yudong and Zheng, Miao and Zhao, Qian and Liu, Kuikun and Zhang, Wenwei and Luo, Ping and Chen, Kai},
  journal={arXiv preprint arXiv:2305.04790},
  year={2023}
}

@inproceedings{goyal2017making,
  title={Making the v in vqa matter: Elevating the role of image understanding in visual question answering},
  author={Goyal, Yash and Khot, Tejas and Summers-Stay, Douglas and Batra, Dhruv and Parikh, Devi},
  booktitle={Proceedings of the IEEE conference on computer vision and pattern recognition},
  pages={6904--6913},
  year={2017}
}

@inproceedings{goyal2017something,
  title={The" something something" video database for learning and evaluating visual common sense},
  author={Goyal, Raghav and Ebrahimi Kahou, Samira and Michalski, Vincent and Materzynska, Joanna and Westphal, Susanne and Kim, Heuna and Haenel, Valentin and Fruend, Ingo and Yianilos, Peter and Mueller-Freitag, Moritz and others},
  booktitle={Proceedings of the IEEE international conference on computer vision},
  pages={5842--5850},
  year={2017}
}

@inproceedings{greff2022kubric,
  title={Kubric: A scalable dataset generator},
  author={Greff, Klaus and Belletti, Francois and Beyer, Lucas and Doersch, Carl and Du, Yilun and Duckworth, Daniel and Fleet, David J and Gnanapragasam, Dan and Golemo, Florian and Herrmann, Charles and others},
  booktitle={Proceedings of the IEEE/CVF conference on computer vision and pattern recognition},
  pages={3749--3761},
  year={2022}
}

@inproceedings{hendricks2021probing,
  title={Probing Image-Language Transformers for Verb Understanding},
  author={Hendricks, Lisa Anne and Nematzadeh, Aida},
  booktitle={Findings of the Association for Computational Linguistics: ACL-IJCNLP 2021},
  pages={3635--3644},
  year={2021}
}

@article{hu2021lora,
  title={Lora: Low-rank adaptation of large language models},
  author={Hu, Edward J and Shen, Yelong and Wallis, Phillip and Allen-Zhu, Zeyuan and Li, Yuanzhi and Wang, Shean and Wang, Lu and Chen, Weizhu},
  journal={arXiv preprint arXiv:2106.09685},
  year={2021}
}

@inproceedings{hu2019evaluating,
  title={Evaluating text-to-image matching using binary image selection (bison)},
  author={Hu, Hexiang and Misra, Ishan and Van Der Maaten, Laurens},
  booktitle={Proceedings of the IEEE/CVF International Conference on Computer Vision Workshops},
  pages={0--0},
  year={2019}
}

@inproceedings{hudson2019gqa,
  title={Gqa: A new dataset for real-world visual reasoning and compositional question answering},
  author={Hudson, Drew A and Manning, Christopher D},
  booktitle={Proceedings of the IEEE/CVF conference on computer vision and pattern recognition},
  pages={6700--6709},
  year={2019}
}

@article{huang2023sparkles,
  title={Sparkles: Unlocking chats across multiple images for multimodal instruction-following models},
  author={Huang, Yupan and Meng, Zaiqiao and Liu, Fangyu and Su, Yixuan and Collier, Nigel and Lu, Yutong},
  journal={arXiv preprint arXiv:2308.16463},
  year={2023}
}

@inproceedings{jhamtani2018learning,
  title={Learning to Describe Differences Between Pairs of Similar Images},
  author={Jhamtani, Harsh and Berg-Kirkpatrick, Taylor},
  booktitle={Proceedings of the 2018 Conference on Empirical Methods in Natural Language Processing},
  pages={4024--4034},
  year={2018}
}

@inproceedings{ji2020action,
  title={Action genome: Actions as compositions of spatio-temporal scene graphs},
  author={Ji, Jingwei and Krishna, Ranjay and Fei-Fei, Li and Niebles, Juan Carlos},
  booktitle={Proceedings of the IEEE/CVF Conference on Computer Vision and Pattern Recognition},
  pages={10236--10247},
  year={2020}
}

@article{jiang2024mixtral,
  title={Mixtral of experts},
  author={Jiang, Albert Q and Sablayrolles, Alexandre and Roux, Antoine and Mensch, Arthur and Savary, Blanche and Bamford, Chris and Chaplot, Devendra Singh and Casas, Diego de las and Hanna, Emma Bou and Bressand, Florian and others},
  journal={arXiv preprint arXiv:2401.04088},
  year={2024}
}

@article{krishna2017visual,
  title={Visual genome: Connecting language and vision using crowdsourced dense image annotations},
  author={Krishna, Ranjay and Zhu, Yuke and Groth, Oliver and Johnson, Justin and Hata, Kenji and Kravitz, Joshua and Chen, Stephanie and Kalantidis, Yannis and Li, Li-Jia and Shamma, David A and others},
  journal={International journal of computer vision},
  volume={123},
  pages={32--73},
  year={2017},
  publisher={Springer}
}

@inproceedings{kuehne2014language,
  title={The language of actions: Recovering the syntax and semantics of goal-directed human activities},
  author={Kuehne, Hilde and Arslan, Ali and Serre, Thomas},
  booktitle={Proceedings of the IEEE conference on computer vision and pattern recognition},
  pages={780--787},
  year={2014}
}

@article{kuznetsova2020open,
  title={The open images dataset v4: Unified image classification, object detection, and visual relationship detection at scale},
  author={Kuznetsova, Alina and Rom, Hassan and Alldrin, Neil and Uijlings, Jasper and Krasin, Ivan and Pont-Tuset, Jordi and Kamali, Shahab and Popov, Stefan and Malloci, Matteo and Kolesnikov, Alexander and others},
  journal={International journal of computer vision},
  volume={128},
  number={7},
  pages={1956--1981},
  year={2020},
  publisher={Springer}
}

@article{li2023mimic,
  title={Mimic-it: Multi-modal in-context instruction tuning},
  author={Li, Bo and Zhang, Yuanhan and Chen, Liangyu and Wang, Jinghao and Pu, Fanyi and Yang, Jingkang and Li, Chunyuan and Liu, Ziwei},
  journal={arXiv preprint arXiv:2306.05425},
  year={2023}
}

@article{li2023otter,
  title={Otter: A Multi-Modal Model with In-Context Instruction Tuning},
  author={Li, Bo and Zhang, Yuanhan and Chen, Liangyu and Wang, Jinghao and Yang, Jingkang and Liu, Ziwei},
  journal={arXiv preprint arXiv:2305.03726},
  year={2023}
}

@article{li2023seed,
  title={Seed-bench: Benchmarking multimodal llms with generative comprehension},
  author={Li, Bohao and Wang, Rui and Wang, Guangzhi and Ge, Yuying and Ge, Yixiao and Shan, Ying},
  journal={arXiv preprint arXiv:2307.16125},
  year={2023}
}

@article{li2023seed2,
  title={Seed-bench-2: Benchmarking multimodal large language models},
  author={Li, Bohao and Ge, Yuying and Ge, Yixiao and Wang, Guangzhi and Wang, Rui and Zhang, Ruimao and Shan, Ying},
  journal={arXiv preprint arXiv:2311.17092},
  year={2023}
}

@inproceedings{lin2014microsoft,
  title={Microsoft coco: Common objects in context},
  author={Lin, Tsung-Yi and Maire, Michael and Belongie, Serge and Hays, James and Perona, Pietro and Ramanan, Deva and Doll{\'a}r, Piotr and Zitnick, C Lawrence},
  booktitle={Computer Vision--ECCV 2014: 13th European Conference, Zurich, Switzerland, September 6-12, 2014, Proceedings, Part V 13},
  pages={740--755},
  year={2014},
  organization={Springer}
}

@inproceedings{pont2020connecting,
  title={Connecting vision and language with localized narratives},
  author={Pont-Tuset, Jordi and Uijlings, Jasper and Changpinyo, Soravit and Soricut, Radu and Ferrari, Vittorio},
  booktitle={Computer Vision--ECCV 2020: 16th European Conference, Glasgow, UK, August 23--28, 2020, Proceedings, Part V 16},
  pages={647--664},
  year={2020},
  organization={Springer}
}

@article{ray2023cola,
  title={cola: A Benchmark for Compositional Text-to-image Retrieval},
  author={Ray, Arijit and Radenovic, Filip and Dubey, Abhimanyu and Plummer, Bryan and Krishna, Ranjay and Saenko, Kate},
  journal={Advances in Neural Information Processing Systems},
  volume={36},
  year={2023}
}

@inproceedings{rombach2022high,
  title={High-resolution image synthesis with latent diffusion models},
  author={Rombach, Robin and Blattmann, Andreas and Lorenz, Dominik and Esser, Patrick and Ommer, Bj{\"o}rn},
  booktitle={Proceedings of the IEEE/CVF conference on computer vision and pattern recognition},
  pages={10684--10695},
  year={2022}
}

@inproceedings{sun2023generative,
  title={Generative multimodal models are in-context learners},
  author={Sun, Quan and Cui, Yufeng and Zhang, Xiaosong and Zhang, Fan and Yu, Qiying and Luo, Zhengxiong and Wang, Yueze and Rao, Yongming and Liu, Jingjing and Huang, Tiejun and others},
  booktitle={CVPR},
  year={2024}
}

@inproceedings{suhr2019corpus,
  title={A Corpus for Reasoning about Natural Language Grounded in Photographs},
  author={Suhr, Alane and Zhou, Stephanie and Zhang, Ally and Zhang, Iris and Bai, Huajun and Artzi, Yoav},
  booktitle={Proceedings of the 57th Annual Meeting of the Association for Computational Linguistics},
  pages={6418--6428},
  year={2019}
}

@article{sun2023eva,
  title={Eva-clip: Improved training techniques for clip at scale},
  author={Sun, Quan and Fang, Yuxin and Wu, Ledell and Wang, Xinlong and Cao, Yue},
  journal={arXiv preprint arXiv:2303.15389},
  year={2023}
}

@misc{team2023internlm,
  title={Internlm: A multilingual language model with progressively enhanced capabilities},
  author={Team, InternLM},
  journal={2023-01-06)[2023-09-27]. https://github. com/InternLM/InternLM},
  year={2023}
}

@inproceedings{wang2023equivariant,
  title={Equivariant similarity for vision-language foundation models},
  author={Wang, Tan and Lin, Kevin and Li, Linjie and Lin, Chung-Ching and Yang, Zhengyuan and Zhang, Hanwang and Liu, Zicheng and Wang, Lijuan},
  booktitle={Proceedings of the IEEE/CVF International Conference on Computer Vision},
  pages={11998--12008},
  year={2023}
}

@inproceedings{wang2022geb+,
  title={Geb+: A benchmark for generic event boundary captioning, grounding and retrieval},
  author={Wang, Yuxuan and Gao, Difei and Yu, Licheng and Lei, Weixian and Feiszli, Matt and Shou, Mike Zheng},
  booktitle={European Conference on Computer Vision},
  pages={709--725},
  year={2022},
  organization={Springer}
}

@article{young2014image,
  title={From image descriptions to visual denotations: New similarity metrics for semantic inference over event descriptions},
  author={Young, Peter and Lai, Alice and Hodosh, Micah and Hockenmaier, Julia},
  journal={Transactions of the Association for Computational Linguistics},
  volume={2},
  pages={67--78},
  year={2014},
  publisher={MIT Press One Rogers Street, Cambridge, MA 02142-1209, USA journals-info~…}
}

@article{zhang2023internlm,
  title={Internlm-xcomposer: A vision-language large model for advanced text-image comprehension and composition},
  author={Zhang, Pan and Wang, Xiaoyi Dong Bin and Cao, Yuhang and Xu, Chao and Ouyang, Linke and Zhao, Zhiyuan and Ding, Shuangrui and Zhang, Songyang and Duan, Haodong and Yan, Hang and others},
  journal={arXiv preprint arXiv:2309.15112},
  year={2023}
}

@article{zhang2023llama,
  title={Llama-adapter: Efficient fine-tuning of language models with zero-init attention},
  author={Zhang, Renrui and Han, Jiaming and Liu, Chris and Gao, Peng and Zhou, Aojun and Hu, Xiangfei and Yan, Shilin and Lu, Pan and Li, Hongsheng and Qiao, Yu},
  journal={arXiv preprint arXiv:2303.16199},
  year={2023}
}

@article{zhang2023llavar,
  title={Llavar: Enhanced visual instruction tuning for text-rich image understanding},
  author={Zhang, Yanzhe and Zhang, Ruiyi and Gu, Jiuxiang and Zhou, Yufan and Lipka, Nedim and Yang, Diyi and Sun, Tong},
  journal={arXiv preprint arXiv:2306.17107},
  year={2023}
}

@article{zhao2023svit,
  title={Svit: Scaling up visual instruction tuning},
  author={Zhao, Bo and Wu, Boya and Huang, Tiejun},
  journal={arXiv preprint arXiv:2307.04087},
  year={2023}
}

@article{zheng2023judging,
  title={Judging llm-as-a-judge with mt-bench and chatbot arena},
  author={Zheng, Lianmin and Chiang, Wei-Lin and Sheng, Ying and Zhuang, Siyuan and Wu, Zhanghao and Zhuang, Yonghao and Lin, Zi and Li, Zhuohan and Li, Dacheng and Xing, Eric and others},
  journal={Advances in Neural Information Processing Systems},
  volume={36},
  year={2023}
}

@inproceedings{zhao2024mmicl,
  title={MMICL: Empowering Vision-language Model with Multi-Modal In-Context Learning},
  author={Zhao, Haozhe and Cai, Zefan and Si, Shuzheng and Ma, Xiaojian and An, Kaikai and Chen, Liang and Liu, Zixuan and Wang, Sheng and Han, Wenjuan and Chang, Baobao},
  booktitle={The Twelfth International Conference on Learning Representations},
  year={2024}
}

@article{zhou2019semantic,
  title={Semantic understanding of scenes through the ade20k dataset},
  author={Zhou, Bolei and Zhao, Hang and Puig, Xavier and Xiao, Tete and Fidler, Sanja and Barriuso, Adela and Torralba, Antonio},
  journal={International Journal of Computer Vision},
  volume={127},
  pages={302--321},
  year={2019},
  publisher={Springer}
}

@inproceedings{zhou2018towards,
  title={Towards automatic learning of procedures from web instructional videos},
  author={Zhou, Luowei and Xu, Chenliang and Corso, Jason},
  booktitle={Proceedings of the AAAI Conference on Artificial Intelligence},
  volume={32},
  number={1},
  year={2018}
}

@misc{bommasani2022opportunities,
      title={On the Opportunities and Risks of Foundation Models}, 
      author={Rishi Bommasani and Drew A. Hudson and Ehsan Adeli and Percy Liang and et al.},
      year={2022},
      eprint={2108.07258},
      archivePrefix={arXiv},
      primaryClass={cs.LG}
}

@misc{openai2024gpt4,
      title={GPT-4 Technical Report}, 
      author={OpenAI et al.},
      year={2024},
      eprint={2303.08774},
      archivePrefix={arXiv},
      primaryClass={cs.CL}
}

@misc{geminiteam2024gemini,
      title={Gemini: A Family of Highly Capable Multimodal Models}, 
      author={Gemini Team et al.},
      year={2024},
      eprint={2312.11805},
      archivePrefix={arXiv},
      primaryClass={cs.CL}
}

@misc{claude,
      title={The Claude 3 Model Family: Opus, Sonnet, Haiku}, 
      author={Anthropic},
      year={2024},
      eprint={2312.11805},
      archivePrefix={arXiv},
      primaryClass={cs.CL}
}

@article{internlmxcomposer2,
      title={InternLM-XComposer2: Mastering Free-form Text-Image Composition and Comprehension in Vision-Language Large Model},
      author={Xiaoyi Dong and Pan Zhang and Yuhang Zang and Yuhang Cao and Bin Wang and Linke Ouyang and Xilin Wei and Songyang Zhang and Haodong Duan and Maosong Cao and Wenwei Zhang and Yining Li and Hang Yan and Yang Gao and Xinyue Zhang and Wei Li and Jingwen Li and Kai Chen and Conghui He and Xingcheng Zhang and Yu Qiao and Dahua Lin and Jiaqi Wang},
      journal={arXiv preprint arXiv:2401.16420},
      year={2024}
}

@article{zhu2023minigpt,
  title={MiniGPT-4: Enhancing Vision-Language Understanding with Advanced Large Language Models},
  author={Zhu, Deyao and Chen, Jun and Shen, Xiaoqian and Li, Xiang and Elhoseiny, Mohamed},
  journal={arXiv preprint arXiv:2304.10592},
  year={2023}
}

@article{Emu2,
    title={Generative Multimodal Models are In-Context Learners}, 
    author={Quan Sun and Yufeng Cui and Xiaosong Zhang and Fan Zhang and Qiying Yu and Zhengxiong Luo and Yueze Wang and Yongming Rao and Jingjing Liu and Tiejun Huang and Xinlong Wang},
    publisher={arXiv preprint arXiv:2312.13286},
    year={2023},
}

@misc{yang2023dawn,
      title={The Dawn of LMMs: Preliminary Explorations with GPT-4V(ision)}, 
      author={Zhengyuan Yang and Linjie Li and Kevin Lin and Jianfeng Wang and Chung-Ching Lin and Zicheng Liu and Lijuan Wang},
      year={2023},
      eprint={2309.17421},
      archivePrefix={arXiv},
      primaryClass={cs.CV}
}

@article{llama3,
  title={Llama 3 Model Card},
  author={AI@Meta},
  year={2024},
  url = {https://github.com/meta-llama/llama3/blob/main/MODEL_CARD.md}
}
}


\end{document}